\documentclass{article}

\usepackage[preprint]{neurips_2026}

\usepackage[utf8]{inputenc} 
\usepackage[T1]{fontenc}    
\usepackage{hyperref}       
\usepackage{url}            
\usepackage{booktabs}       
\usepackage{amsfonts}       
\usepackage{nicefrac}       
\usepackage{microtype}      
\usepackage{xcolor}         
\usepackage{amsmath}

\usepackage{graphicx}
\usepackage{float}
\usepackage{wrapfig}

\usepackage{xcolor}
\definecolor{linkteal}{HTML}{237A68}
\definecolor{purple}{HTML}{6B4C93}
\usepackage[table]{xcolor}
\definecolor{sectionfill}{RGB}{242,239,232}

\hypersetup{
    colorlinks=true,
    linkcolor=linkteal,  
    citecolor=linkteal,  
    urlcolor=linkteal    
}

\title{On-the-fly Weight Generation: \ \ \ \ \ \ \ \ \ \ \ \ \ \ \ \ \ \ \ \ \ \ \ \ \ \ \ \ \ \ \ \ \ \ \ \ \ \ \ \ \ \ A Hypernetwork Proof of Concept on ARC-1D}

\author{%
  Fabio J.~Fehr \\
  University of Oxford\\
  \And
Philip Torr \\
University of Oxford
}

\begin{document}

\maketitle
\begin{center}
  \vspace{-2.5em}
\small
\href{https://github.com/FJFehr/on-the-fly-models}{%
    \raisebox{-0.2em}{\includegraphics[height=1.2em]{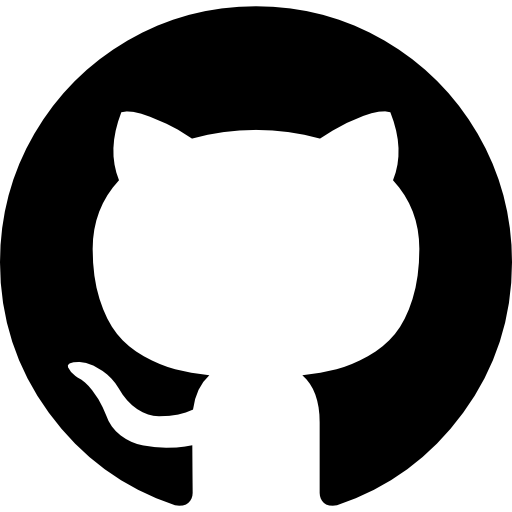}}\ Code
}
\end{center}

\begin{abstract}
General-purpose models can adapt to many tasks from context, while specialised models can execute individual functions with less capacity. Yet obtaining such specialists requires task-specific training or adaptation. We ask whether they can instead be generated directly from a few demonstrations. Using ARC-1D as a controlled testbed, we show that individual transformations can be represented by tiny specialist models, and that a hypernetwork can generate their parameters from context. The generated parameters form a structured weight space, while the resulting specialists show partial compositional generalisation and generalisation to transformations not seen during training. In both settings, removing explicit task identifiers improves generalisation beyond the training transformations. Together, these results provide a proof of concept that few-shot task context can be compiled on-the-fly into compact executable model parameters, and that the resulting weight space can support reuse and generalisation beyond known functions.
\end{abstract}












\section{Introduction}

In-context learning has shown that large general-purpose models can infer task-specific behaviour from examples at inference time, without explicit gradient-based adaptation \citep{brown2020_fewshotlearners,min-etal-2022-rethinking,dong-etal-2024-survey}. At the same time, specialised models can execute individual functions with substantially less capacity, but typically require task-specific training, fine-tuning, or distillation \citep{hsieh-etal-2023-distilling,Wang2020,tang2019distillingtaskspecificknowledgebert}. This suggests an alternative form of inference-time adaptation: rather than retaining the inferred behaviour within a general-purpose model, it could be instantiated as a lightweight task-specific model and executed independently. This perspective also extends the paradigm of tool use in language-model agents \citep{schick2023_toolformer,jin2025searchr,qian2025toolrl} by allowing a model to construct a specialised model rather than select a predefined tool or generate code. This leads to our central question: \textit{can compact task-specific models be constructed on-the-fly from only a few demonstrations?}

More generally, this can be viewed as a problem of few-shot function induction. A few demonstrations specify a function that must be inferred and applied to new inputs. While this function can remain implicit in context or be expressed through programs or latent representations, we instead consider representing it directly as model parameters. Hypernetworks provide one mechanism for doing so by mapping contextual information to model weights \citep{ha2017hypernetworks,Chauhan_2024}. We study whether demonstrations can be used to instantiate the complete parameters of a compact standalone network, separating \emph{task inference} from \emph{task execution}.



By representing functions as model parameters, we can study whether weight space reflects structure in the functions themselves. Recent work has modelled neural-network weights as a generative domain \citep{han2026surveyweightspacelearning,zeng2026generative,wang2026position}, while meta-learning and abstract reasoning have explored structured parameter, program, and function spaces that support adaptation and composition \citep{rusu2018metalearning,macfarlane2025searching,shirakami2026compositional}. A related challenge is whether this can be achieved without an explicit task identifier: tiny-model approaches to ARC-AGI often use task-specific conditioning \citep{wang2025hierarchicalreasoningmodel,jolicoeur2025less,gao2025universalreasoningmodel}, while larger-scale text-to-LoRA methods condition generated adaptations on explicit task descriptions \citep{charakorn2025texttolora}. These considerations lead to two questions: do related functions occupy related regions of weight space that support reuse and generalisation, and can the underlying function be inferred from support examples alone, without an explicit task identifier?

We study these questions on ARC-1D \citep{xu2024llms}, a controlled setting at the intersection of abstract reasoning and sequence modelling. Like ARC-AGI \citep{chollet2019measure,chollet2026arcprize2025technical,arcagi3_2026}, ARC-1D requires inferring an unknown transformation from a few input--output demonstrations, but operates over one-dimensional discrete sequences that can be modelled with compact sequence architectures. This makes ARC-1D a useful testbed for studying few-shot function induction and the generation, reuse, and generalisation of compact executable models. Our goal is not to build a specialised ARC-1D solver, but to study whether demonstrations can be converted on-the-fly into compact model parameters. Our \textbf{contributions} are as follows:

\begin{itemize}
\item \textbf{Specialisation reduces the capacity required for task execution.} Tiny models accurately represent individual ARC-1D transformations, while the same architectures degrade when trained jointly across transformations, even with access to task identifier (Section~\ref{sec:target_capacity}).
\item \textbf{Hypernetworks can generate compact specialists directly from demonstrations.} The generated models separate task inference from execution and form a structured weight space organised by transformation, with related functions occupying nearby regions (Section~\ref{sec:hypernetwork_generation}).
%
\item \textbf{Generated specialists encode reusable functions and generalise beyond known transformations.} Once generated from a support set, a specialist can be decoupled from the hypernetwork and reused on new instances of the same transformation. Removing task identity further improves transfer to novel compositions of known transformations and to transformations not seen during training (Section~\ref{sec:generalisation}).
\end{itemize}

\begin{figure}[t]
    \centering
    \includegraphics[width=\linewidth]{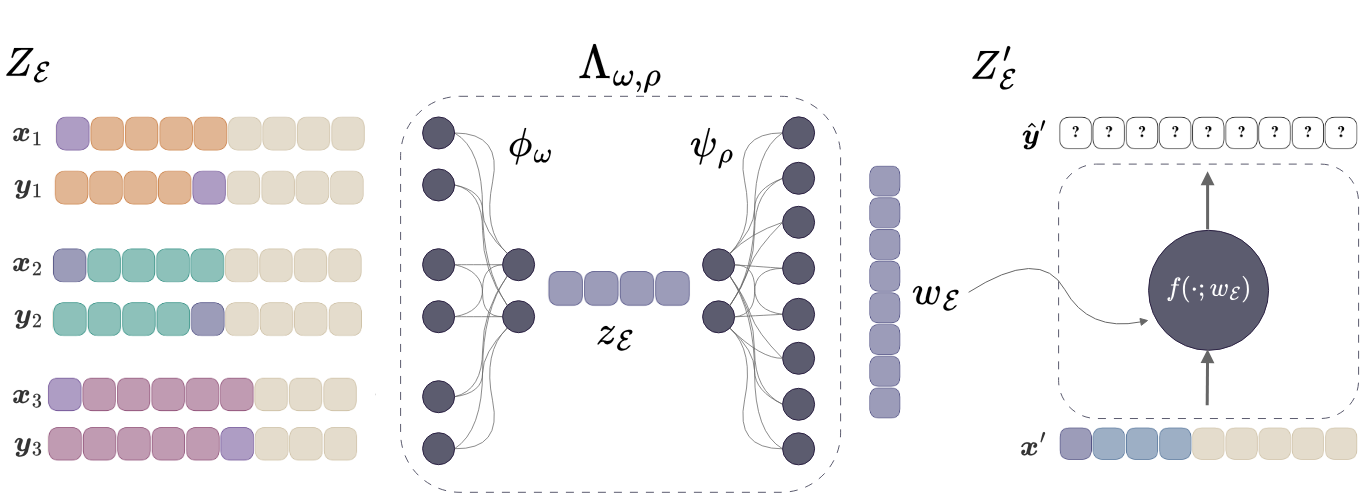}
    \caption{\textbf{Hypernetwork weight generation.}
The support set $Z_{\mathcal E}=\{(x_i,y_i)\}_{i=1}^{K}$ is encoded by the task encoder $\phi_\omega$ into an episode representation $z_{\mathcal E}$, which the weight decoder $\psi_\rho$ maps to the complete parameters $w_{\mathcal E}$ of the episode-specific predictor $f(\cdot;w_{\mathcal E})$. Once generated, the predictor operates independently of the support set, receiving only a new input $x'$ to produce its output.}
    \label{fig:hypermodel}
    \vspace{-1em}
\end{figure}

\section{Related Work}

\paragraph{Small and Specialised Models.}
Large general-purpose models provide broad capabilities, while smaller task-specific models can achieve strong performance on narrower tasks with substantially less capacity~\citep{hsieh-etal-2023-distilling}. Such specialists are commonly obtained through fine-tuning or distillation~\citep{Wang2020,tang2019distillingtaskspecificknowledgebert}, while lottery-ticket results suggest that much smaller subnetworks can retain the functionality of larger networks~\citep{malach2020_lotteryticket}. Recent abstract-reasoning models, including HRM~\citep{wang2025hierarchicalreasoningmodel}, TRM~\citep{jolicoeur2025less}, and URM~\citep{gao2025universalreasoningmodel}, further show that small networks can solve complex reasoning tasks through iterative computation, extended through stochastic exploration~\citep{sghaier2026probabilistictinyrecursivemodel} and adaptive fixed-point iteration~\citep{movahedi2026fixedpoint}. Similarly, neural algorithmic reasoning has progressed from specialist models executing individual algorithms towards generalist models learning across collections of algorithms~\citep{ibarz2022a}. Rather than training each specialist independently or representing all functions within shared parameters, we ask whether a general model can infer a function from a few demonstrations and directly generate the corresponding compact specialist.

\paragraph{Hypernetworks and Weight Generation.}
Hypernetworks generate the parameters of a target network with a forward pass, enabling rapid instantiation of task-specific weights ~\citep{ha2017hypernetworks,Chauhan_2024}. Subsequent work has improved scalability through weight chunking~\citep{Zhao_Kobayashi_Sacramento_von_Oswald_2020}, low-rank factorisations~\citep{Bertinetto2016,hypertransformer2022}, and parameter-efficient generation, including adapters and LoRA updates~\citep{karimi-mahabadi-etal-2021-parameter,ruiz2024,charakorn2025texttolora,liang2025draganddrop}. In parallel, generative weight-space models learn distributions over neural-network parameters~\citep{han2026surveyweightspacelearning,zeng2026generative,wang2026position}, using conditional diffusion to generate task- or domain-specific parameters~\citep{jin2024conditionalloraparametergeneration,xie2024weight}, recurrent diffusion to scale to complete large networks~\citep{wang2025scaling}, and flow matching for efficient full-network generation~\citep{gupta2026deepweightflow}. Rather than generating from collections of trained parameters or explicit task conditions, we infer a function from a few input--output demonstrations and instantiate it as the complete weights of a small executable model. We further ask whether generating many such models induces a structured weight space that supports generalisation between functions.

\paragraph{Abstract Reasoning and Function Induction.}

ARC-AGI ~\citep{chollet2019measure} frames abstract reasoning as few-shot function induction, where an unknown transformation must be inferred from input--output demonstrations and applied to a held-out query. ARC-1D translates this setting to one-dimensional discrete sequences, providing a controlled setting for studying function induction with sequence models~\citep{xu2024llms}. Approaches to ARC style problems differ in how this inferred function is represented and recovered, ranging from direct prediction to explicit or latent program induction~\citep{li2025combining}. Latent Program Networks, for example, learn a latent program space and search within it at test time~\citep{macfarlane2025searching}, while recent work on ARC-1D learns structured latent function spaces that support compositional generalisation~\citep{shirakami2026compositional}. Related work has proposed puzzle-specific weight perturbations and suggests that hypernetworks mapping puzzle embeddings to model parameters as a direction for ARC~\citep{liao2025arcagipretraining}. Similarly, HRM conditions predictions on learned puzzle-ID embeddings rather than few-shot context; subsequent analysis identifies this as a major limitation because inference is restricted to puzzle identities observed during training, motivating approaches that infer task identity directly from demonstrations \citep{schurholt2025hrm}.
We instead amortise function induction into a hypernetwork that maps demonstrations directly to the complete parameters of a reusable specialist, and ask whether task identity can be inferred from these demonstrations without an explicit puzzle embedding. We further study whether the resulting weight space supports generalisation to novel compositions and transformations not seen during training.


\paragraph{Meta-Learning and Few-Shot Adaptation.}
Our formulation is closely related to meta-learning, where models learn across tasks to adapt from few examples~\citep{Hospedales2022}. Gradient-based methods such as MAML~\citep{maml2017} and Reptile~\citep{nichol2018firstordermetalearningalgorithms} learn shared parameters that can be adapted to each task through a small number of optimisation steps, while LEO ~\citep{rusu2018metalearning} performs this task-specific optimisation in a learned low-dimensional latent parameter space. Other approaches avoid task-specific optimisation through differentiable solvers~\citep{bertinetto2018metalearning}, fast weights~\citep{Munkhdalai2017}, or direct prediction of task-specific parameters~\citep{Bertinetto2016,nava2022metalearning}. 
Our setting differs from conventional few-shot meta-learning because augmentation provides many training episodes for each transformation; the few-shot constraint applies within each episode, where a small support set must specify the function and the parameters that execute it. This follows the data-augmented training regime common to small ARC models.


\section{Methodology}
\label{sec:methodology}

\subsection{Episodic Meta-Learning Formulation}

We formulate weight generation as episodic meta-learning. Let
$\mathbb{E}$ denote a collection of episodes, where each
$\mathcal{E}\in\mathbb{E}$ contains a support set and a query set,
\begin{equation*}
    Z_{\mathcal E}=\{(x_i,y_i)\}_{i=1}^{K},
    \qquad
    Z'_{\mathcal E}=\{(x'_i,y'_i)\}_{i=1}^{Q}.
\end{equation*}
The support and query examples share the same underlying transformation.
For ARC-1D, $K=3$ and $Q=1$, with inputs and outputs given by
variable-length sequences $x,y\in\mathcal V^*$ over
$\mathcal V=\{0,\ldots,9\}$. At test time, the model observes
$Z_{\mathcal E}$ and the query input $x'$, and must predict its output $y'$.

At the level of an individual episode, a base learner $\Lambda$ maps the
support examples to episode-specific parameters
$w_{\mathcal E}\in\mathbb R^p$,
\begin{equation}
    w_{\mathcal E}=\Lambda(Z_{\mathcal E}),
    \qquad
    \hat y'=f(x';w_{\mathcal E}),
    \label{eq:episodic-learning}
\end{equation}
where $f(\cdot;w_{\mathcal E})$ is the episode-specific predictor.
Meta-learning learns across episodes how to construct these predictors.
In our formulation, $\Lambda$ is a learned feed-forward weight generator
that produces the complete parameters of $f$ directly from the support
examples, without task-specific optimisation. The query is therefore
held out from the construction of $w_{\mathcal E}$, while its target is
used to supervise the weight generator during meta-training.

\subsection{Hypernetwork Weight Generation}

We implement the base learner $\Lambda$ as an encoder--decoder
hypernetwork (Figure~\ref{fig:hypermodel}). A task encoder
$\phi_\omega$ compresses the support set into an episode representation
$z_{\mathcal E}$, and a weight decoder $\psi_\rho$ maps this representation
to the complete parameters of the episode-specific predictor:
\begin{equation}
    z_{\mathcal E}=\phi_\omega(Z_{\mathcal E}),
    \qquad
    w_{\mathcal E}=\psi_\rho(z_{\mathcal E}),
    \qquad
    \Lambda_{\omega,\rho}
    =
    \psi_\rho\circ\phi_\omega.
    \label{eq:weight-generation}
\end{equation}
Task inference and task execution are therefore separated: the support
examples are processed by the hypernetwork, while the generated parameters
$w_{\mathcal E}$ provide the interface to the predictor that executes the
inferred transformation.

\paragraph{Task encoder.}
The encoder $\phi_\omega$ is a Transformer operating on the $K$ support
pairs. Each token is represented as
\begin{equation*}
    e(t)
    =
    e_{\mathrm{tok}}(t)
    +
    e_{\mathrm{ex}}(i)
    +
    e_{\mathrm{role}}(t),
\end{equation*}
combining token, support-example, and input/output-role embeddings.
RoPE~\citep{su_2024_rope} provides positional information, with positions
reset for each input and output sequence. Learned-query attention pooling
reduces the resulting support representations to a single episode
representation
$z_{\mathcal E}\in\mathbb R^{d_h}$.

\paragraph{Weight decoder.}
In the task-ID-conditioned variant, we augment the support-derived
representation as $\tilde z_{\mathcal E} = z_{\mathcal E} W_{\mathrm{id}}c_{\mathcal E}$, 
where $c_{\mathcal E}$ is a one-hot transformation identifier and
$W_{\mathrm{id}}\in\mathbb R^{d_h\times14}$ is randomly initialised and
kept fixed during training. Without task-ID conditioning, no additional
information is provided and
$\tilde z_{\mathcal E}=z_{\mathcal E}$. The decoder $\psi_\rho$ is a two-layer MLP without a bias term,
\begin{equation}
    z'_{\mathcal E}
    =
    \mathrm{GELU}(W_1\tilde z_{\mathcal E}),
    \qquad
    w_{\mathcal E}
    =
    W_2z'_{\mathcal E},
    \label{eq:weight-decoder}
\end{equation}
with
$W_1\in\mathbb R^{d'\times d_h}$,
$W_2\in\mathbb R^{p\times d'}$,
and $d'\ll p$.
The intermediate representation $z'_{\mathcal E}$ therefore provides a
low-dimensional parameter-space bottleneck: although
$w_{\mathcal E}\in\mathbb R^p$ specifies the complete target model,
all generated weight vectors lie in the learned subspace spanned by
$W_2$. Thus, the transformation inferred from the demonstrations must be
expressed through a constrained region of the target model's parameter
space.

\paragraph{Episode-specific predictor.}
The generated vector $w_{\mathcal E}\in\mathbb R^p$ contains all parameters
of $f(\cdot;w_{\mathcal E})$, a three-layer Canon Transformer
\citep{allen-zhu2025physics} with RoPE. This architecture is fixed across
episodes, while its parameters are generated anew from each support set
\footnote{Indeed, generating the architecture and weights jointly would be more expressive, we leave this for future work.}.
Once $w_{\mathcal E}$ has been generated,
$f(\cdot;w_{\mathcal E})$ is a standalone model: it can be separated from
the hypernetwork and applied to new inputs using only its generated
parameters, without access to either $Z_{\mathcal E}$ or the
encoder--decoder that produced them.

\subsection{Meta-Training Objective}

We train the hypernetwork end-to-end across episodes. For each episode, only
the support set $Z_{\mathcal E}$ is provided to the hypernetwork, which
encodes it into an episode representation and decodes it into the
target-model parameters $w_{\mathcal E}$. The generated predictor must then
reconstruct the demonstrated input--output mapping through these parameters.
This gives an autoencoder-like reconstruction task in which task information
is compressed through a low-dimensional  bottleneck and re-expressed
as an executable model. The episode loss is
\begin{equation}
    \mathcal{L}_{\mathcal E}(\omega,\rho)
    =
    \frac{1}{|Z_{\mathcal E}\cup Z'_{\mathcal E}|}
    \sum_{(x,y)\in Z_{\mathcal E}\cup Z'_{\mathcal E}}
    \mathcal{L}_{\mathrm{pred}}
    \left(
        f(x;w_{\mathcal E}),
        y
    \right),
    \label{eq:episode-loss}
\end{equation}
where
$w_{\mathcal E}=\psi_\rho\!\left(\phi_\omega(Z_{\mathcal E})\right)$.
The meta-training objective averages $\mathcal{L}_{\mathcal E}$ across
training episodes.

Unlike the held-out-query objective commonly used in episodic meta-learning,
we supervise on both the support and query examples
$Z_{\mathcal E}\cup Z'_{\mathcal E}$. The support loss requires the generated
parameters to capture the demonstrated function, while the query loss
requires that function to generalise beyond the demonstrations. Because the
entire computation from $Z_{\mathcal E}$ through $w_{\mathcal E}$ to the
predictions of $f$ is differentiable, these losses jointly train the task
encoder $\phi_\omega$ and weight decoder $\psi_\rho$.

 \section{Experiments} 

\subsection{Data}
\label{sec:data}

\begin{figure}[h!]
\vspace{-1em}
    \centering
    \includegraphics[width= \linewidth]{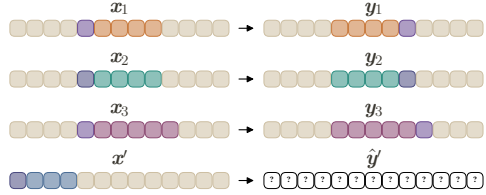}
    \caption{\textbf{Flip transformation: Example ARC-1D episode} The support set $Z_{\mathcal E}$ contains three input--output pairs specifying the transformation, which is applied to the query input in $Z'_{\mathcal E}$. See Appendix~\ref{sec:task_examples} for more examples.}
    \label{fig:arc1d-example}
\end{figure}

\paragraph{ARC-1D.} We evaluate on ARC-1D\footnote{\url{https://github.com/khalil-research/1D-ARC}} \citep{xu2024llms}, a one-dimensional variant of ARC-AGI for studying abstract reasoning from few demonstrations with sequence models. As illustrated in Figure~\ref{fig:arc1d-example}, inputs and outputs are sequences of coloured cells, with transformations involving concepts such as objects, counting, and spatial relations. Each episode $\mathcal E$ contains three support pairs, $Z_{\mathcal E}=\{(x_1,y_1),(x_2,y_2),(x_3,y_3)\}$, and one query pair, $Z'_{\mathcal E}=\{(x',y')\}$, corresponding to $K=3$ and $Q=1$ in Section~\ref{sec:methodology}.

\vspace{-1em}
\paragraph{Transformation subset.}
We use $14$ of the $18$ ARC-1D transformation categories, retaining those that can be executed by a standalone predictor without continued access to the support set. We additionally exclude \texttt{Padded\_Fill} because of its substantially different sequence-length distribution. ARC-1D provides $50$ episodes for each transformation, which we split into $40$ training, $5$ validation, and $5$ test episodes. The same transformation categories therefore occur across all three splits, while the episodes themselves are disjoint. In the task-ID-conditioned variant, the transformation category provides $c_{\mathcal E}$.

\paragraph{Augmentation.}
Following ARC-style augmentation
\citep{wang2025hierarchicalreasoningmodel}, we generate $1K$ augmentations per
base training episode using colour remapping and sequence shifts that preserve
the underlying transformation. Colour mappings are sampled independently
across examples, except for transformations requiring a globally consistent
marker colour, and augmented training episodes matching held-out examples are
discarded to prevent leakage. This yields $561K$ augmented training episodes.
Validation and test episodes are augmented separately using colour remapping
only, with $20$ remappings per held-out episode, yielding $100$ evaluation
episodes per category.

\paragraph{Experimental Setup and Evaluation.}
Full architecture, optimisation, and implementation details are provided in Appendix~\ref{sec:training_details}. We additionally ablate the use of Canon layers~\citep{allen-zhu2025physics} and the Muon optimiser~\citep{liu2025muon} in Appendix~\ref{sec:ablations}, finding that both improve downstream performance. We evaluate query predictions using exact-match accuracy as our primary metric, which requires the complete predicted sequence to match the target. We additionally report token (per-cell) accuracy when analysing partial solutions.


\subsection{Specialised Models Require Less Capacity}
\label{sec:target_capacity}

\begin{wrapfigure}{r}{0.4\textwidth}
\vspace{-1.2em}
    \centering
    \includegraphics[width=\linewidth]{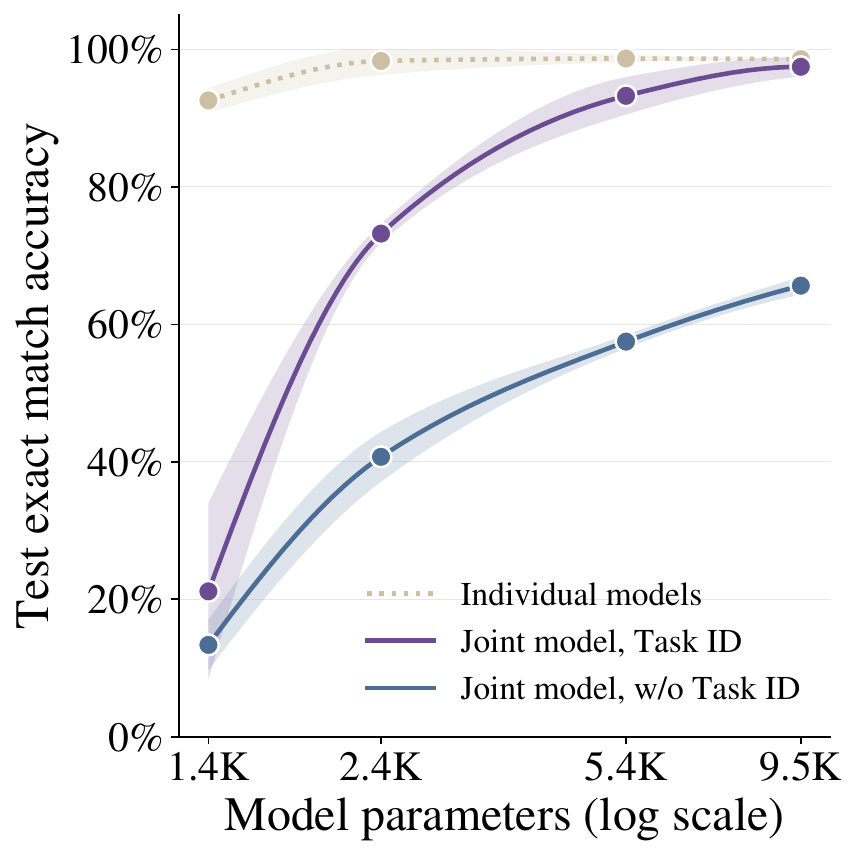}
    \caption{\textbf{Specialisation requires less capacity.}
    Individual models saturate earlier than joint models, even when provided
    with transformation identity. Bands show $\pm1$ s.d. across 5 seeds.}
    \label{fig:capacity_cliff}
\end{wrapfigure}

We first ask how much capacity is required to represent an individual ARC-1D transformation. Recent results show that small recurrent models can solve complex ARC tasks \citep{wang2025hierarchicalreasoningmodel,jolicoeur2025less,gao2025universalreasoningmodel}, motivating us to use a deliberately small target architecture. We progressively simplify a Transformer with RoPE \citep{su_2024_rope} to the three-layer Canon Transformer \citep{allen-zhu2025physics} used throughout our subsequent experiments. Our aim is not to identify an optimal architecture, but to obtain a simple target model whose capacity can be varied directly. Full architecture configurations are provided in Appendix Table~\ref{tab:architectures}.

\begin{figure}[h!]
    \centering
    \includegraphics[width=\linewidth]{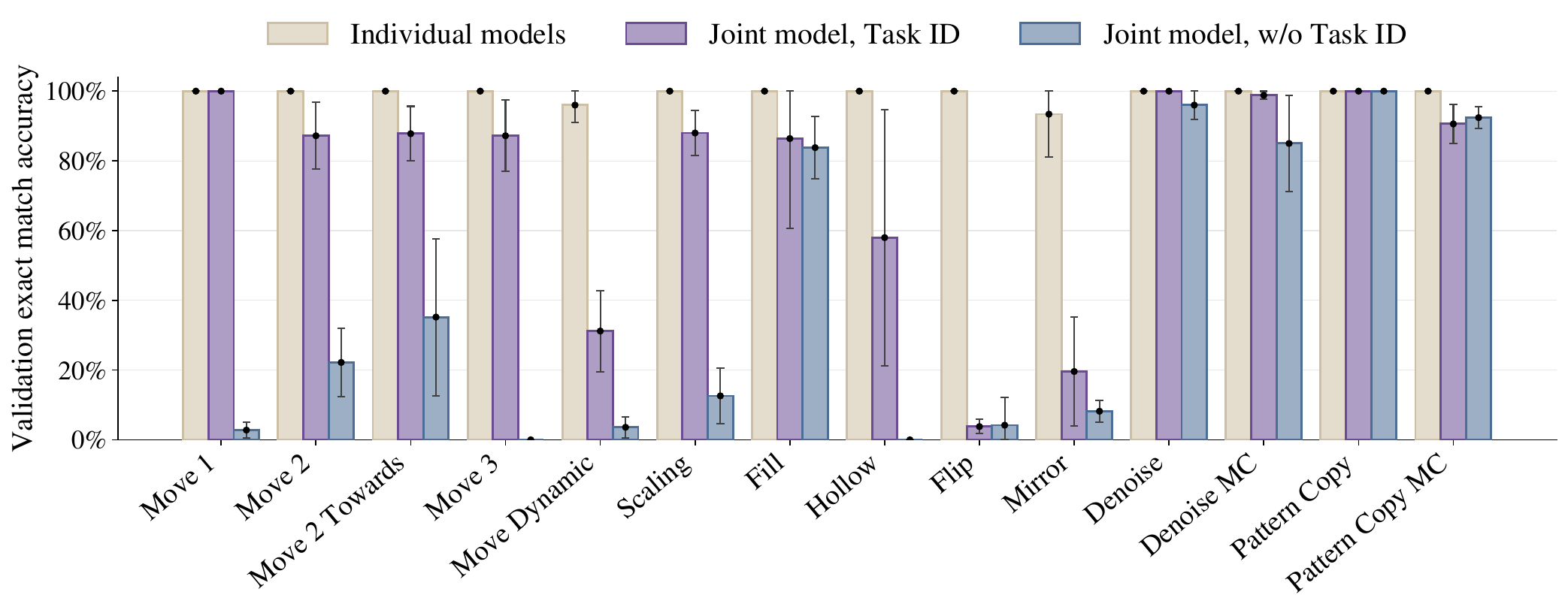}
    \vspace{-1.5em}
    \caption{\textbf{The joint-training penalty is concentrated in a few
    transformations.} Per-transformation exact-match validation accuracy at
    2.4K parameters. Error bars show $\pm1$ s.d. across 5 seeds.}
    \label{fig:per_task}
\end{figure}

Figure~\ref{fig:capacity_cliff} reveals a clear cost to sharing parameters.
Individual models saturate at 98.9\% accuracy with only 2.4K parameters, while
the same sized architecture reaches 74.0\% when trained jointly, even when given the
transformation identity. Increasing capacity closes this gap, suggesting that
individual transformations are cheap to represent but require additional
capacity when stored in shared parameters. This penalty is concentrated in a
few transformations (Figure~\ref{fig:per_task}); most remain near their
individual ceiling, while \textit{Flip}, \textit{Mirror} and \textit{Moves} $1$, $3$ fall from 100\%. We provide figures of the alternative model sizes in Appendix \ref{sec:per_task_plots}.

These results separate the capacity required to represent multiple functions
in shared parameters from that required to execute an individual function.
This motivates separating task inference from execution: rather than storing
all transformations in shared target-model parameters, a hypernetwork can
infer the transformation from the support set and instantiate a small
standalone predictor to execute it. This leads to our central question:
\textit{given that a tiny network can represent each solution, can a
hypernetwork learn to generate those networks from examples?}
\vspace{1em}

\subsection{Generating Specialised Models with Hypernetworks}
\label{sec:hypernetwork_generation}

The previous experiment showed that representing multiple transformations in shared target-model parameters incurs a capacity cost.  We therefore ask whether a hypernetwork can instead generate a small, specialised model for each episode. We train a 10K-parameter hypernetwork $\Lambda_{\omega,\rho}=\psi_\rho\circ\phi_\omega$ jointly across all 14 transformations to map the support set $Z_{\mathcal E}$ to the complete parameters $w_{\mathcal E}$ of the same 1.4K-parameter target architecture used above. We compare generation from the support set alone with a task-ID-conditioned variant that additionally receives a frozen transformation embedding; alternative embedding placements and learned embeddings are considered in Appendix~\ref{sec:ablations}. Full hypernetwork configurations are given in Appendix Table~\ref{tab:architectures}.
 \vspace{-1em}
\begin{figure}[h!]
    \centering
    \includegraphics[width=\linewidth]{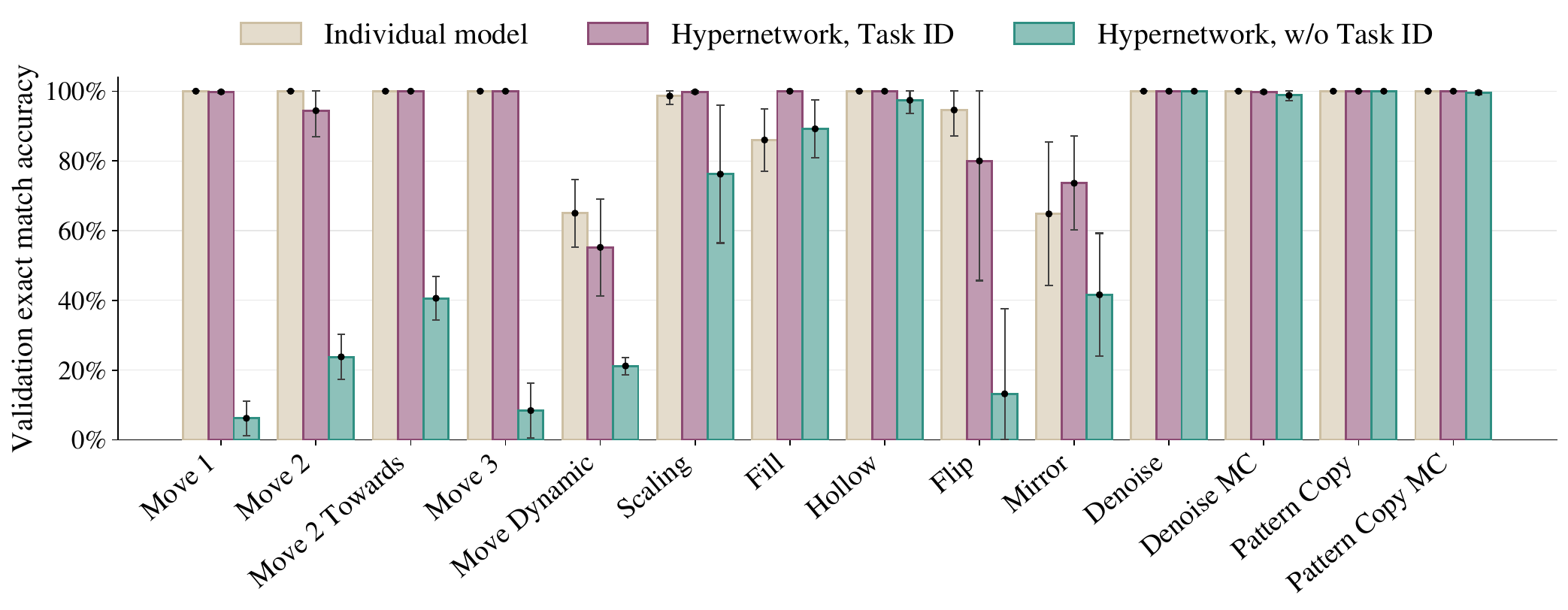}
    \vspace{-1em}
    \caption{\textbf{Generated models recover specialist performance.}
Per-task validation exact-match accuracy for individually trained and
hypernetwork-generated models, with and without task IDs, using the smallest 1.4K-parameter target model. Error bars show $\pm1$ s.d. across 5 seeds.}
    \label{fig:hypernetwork_per_task}
\end{figure}

\begin{figure}[h!]
    \centering
    \includegraphics[width=\linewidth]{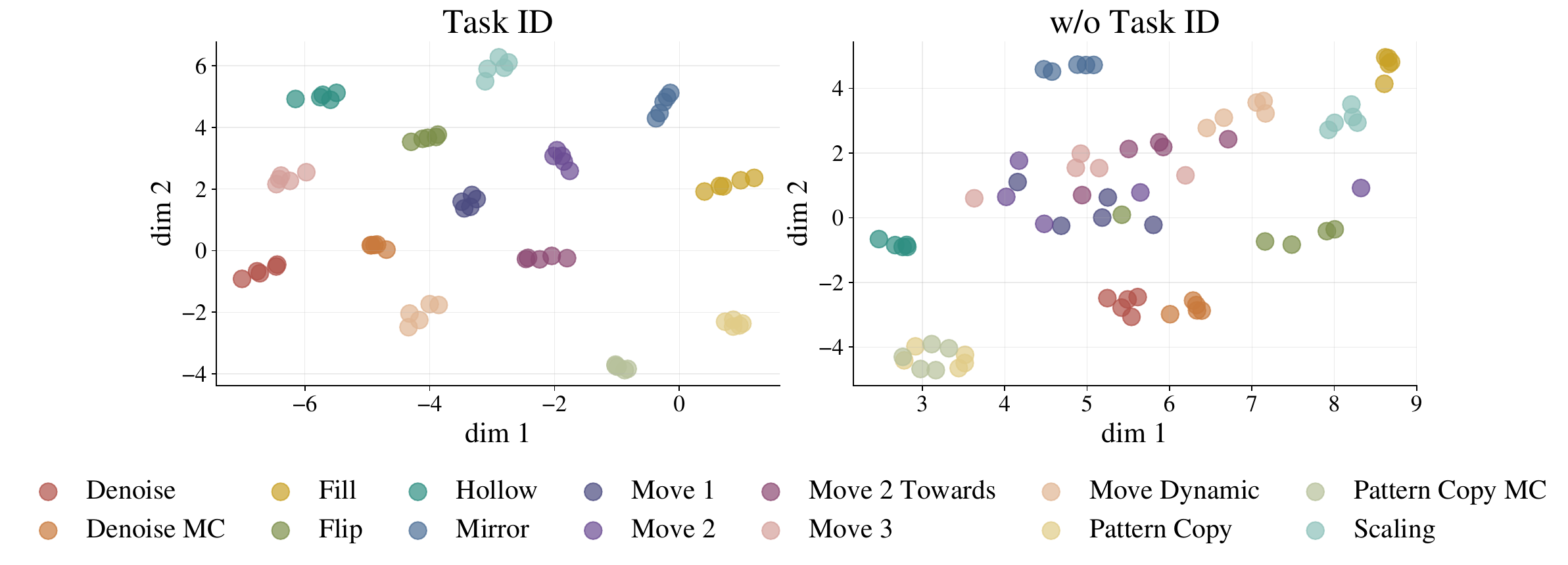}
    \caption{\textbf{Generated parameters form a task-structured weight space.}
t-SNE projection of generated target-model parameters, coloured by transformation.
\textbf{Left:} With a frozen task embedding, tasks are fully separable (100\% linear probe accuracy).
\textbf{Right:} Without task identity, structure emerges from the support examples alone (60.9\% linear probe accuracy). }
    \label{fig:tsne}
\end{figure}

We find that hypernetwork-generated models largely recover the performance of individually trained specialists (Figure~\ref{fig:hypernetwork_per_task}). We also test whether the hypernetwork improves data efficiency by varying the amount of augmented training data, but find no gain over individual models at this scale (Appendix~\ref{sec:data_efficiency}). We next consider whether transformations are reflected in the structure of the generated weight space in Figure~\ref{fig:tsne}. With a frozen task embedding, transformations are perfectly separated (100\% linear-probe accuracy). Without task identity, structure still emerges from the support examples alone (60.9\% linear-probe accuracy), with overlap concentrated among related transformations: the \textit{Move} variants overlap with one another, as do the \textit{Pattern Copy} variants. This suggests that related functions occupy nearby regions of weight space. PCA and UMAP show the same qualitative structure (Appendix Figures~\ref{fig:pca} \& \ref{fig:umap}).

 \vspace{-1em}

Together, these results show that a hypernetwork can generate small task-specific models while learning a structured weight space across transformations. However, all transformations considered so far were observed during training, and the partial overlap between related functions suggests that shared structure may be useful rather than a limitation. We therefore consider whether the hypernetwork can move beyond generating known functions: first by reusing generated models across examples of the same transformation, then by composing related transformations, and finally by generating models for entirely unseen tasks.

\subsection{Towards Generalisation with Hypernetworks}
\label{sec:generalisation}

\begin{wrapfigure}{r}{0.4\textwidth}
\vspace{-3.6em}
    \centering
    \includegraphics[width=\linewidth]{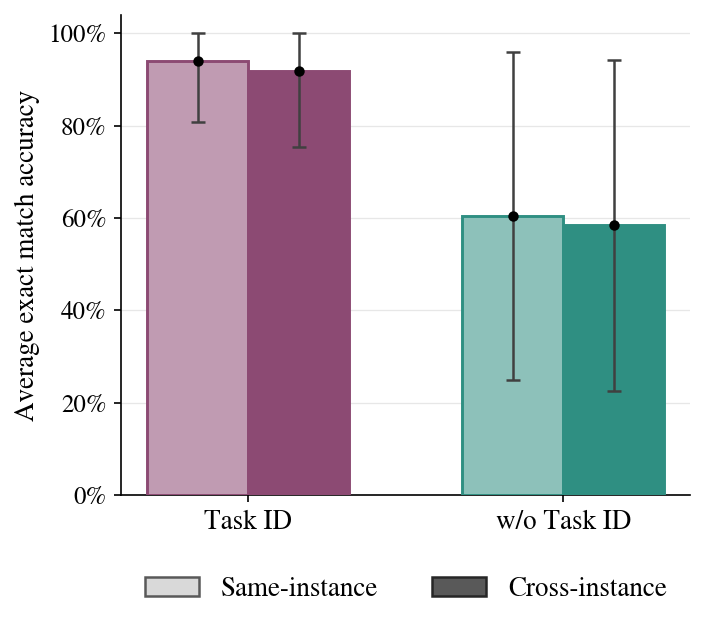}
    \vspace{-0.44em}
    \caption{\textbf{Generated specialists generalise across episodes.}
    Exact-match accuracy on the query paired with the generating support set
    and on queries from other episodes of the same transformation. Bars show
    the macro-average over 14 transformations with $\pm1$ s.d.}
    \vspace{-2em}
    \label{fig:generalization_loo}
\end{wrapfigure}

The previous section showed that hypernetworks can generate effective specialists from demonstrations and that the resulting parameters exhibit structure across transformations. We next ask whether these generated parameters encode functions that generalise beyond the support sets from which they were produced. We consider three settings: \textbf{(1)} reusing generated weights across new episodes of the same transformation, \textbf{(2)} generating weights for novel compositions of transformations observed independently during meta-training, and \textbf{(3)} generating weights for transformations not observed during meta-training.

\paragraph{Generate Once, Execute Many.}
We next ask whether generated parameters encode reusable functions rather
than support-specific solutions. For each transformation, we generate
$w_{\mathcal E}$ from one validation support set and hold it fixed while
evaluating $f(\cdot;w_{\mathcal E})$ on all remaining validation and test
queries, with no further support context, training, or adaptation. As shown in Figure~\ref{fig:generalization_loo}, macro-averaged accuracy remains close to performance on the query paired with the generating support set, both with and without task-ID conditioning. Thus, a generated specialist can be instantiated once and reused across unseen inputs from the same transformation.

\begin{figure}[b!]
    \centering
    \includegraphics[width=\linewidth]{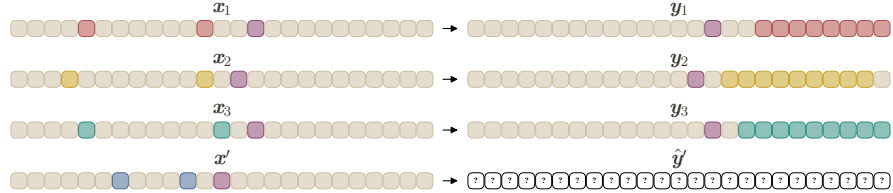}
    \caption{\textbf{Fill $\circ$ Mirror: Example compositional ARC-1D episode} The support pairs specify the composition of Mirror followed by Fill, where both transformations are observed independently during training but their composition is held out. See Appendix~\ref{sec:compositional_examples} for more examples.}
    \label{fig:composition-example}
\end{figure}

\paragraph{Compositional Generalisation.}
We next test whether the generated specialists can transfer to combinations of functions learned independently during meta-training. We construct 10 held-out compositions of known transformations, such that each constituent transformation is observed during training but the composition itself is never seen (Figure~\ref{fig:composition-example}). We denote a composition as $B \circ A$, where $A$ is applied first and $B$ second. We evaluate the same models in-distribution on the 14 transformations used during training and zero-shot on the held-out compositions (Table~\ref{tab:generalisation}). Task-ID conditioning improves in-distribution performance, but this advantage reverses on the compositions: removing the task ID improves transfer, particularly at the token level. Exact-match performance remains near zero for both variants, with the few exact solutions concentrated in compositions involving \texttt{Mirror} and \texttt{Denoise MC} (Appendix Table~\ref{tab:compositional_breakdown}). Thus, explicit task identity improves in-distribution specialisation but appears to constrain the composition of known transformations into new functions.


\paragraph{Unseen Generalisation.}
We finally test transfer to transformations entirely absent from meta-training using leave-one-category-out evaluation: each of the 14 transformations is withheld in turn, with the hypernetwork trained on the remaining 13 and results macro-averaged across held-out transformations (Table~\ref{tab:generalisation}). As with compositional generalisation, task-ID conditioning improves in-distribution performance but hinders transfer, with the support-only model achieving higher token and exact-match accuracy on unseen transformations. Unlike the compositional setting, exact solutions emerge more clearly, suggesting that support examples can provide a transferable description of an unseen function rather than relying solely on task identity. Per-transformation results are reported in Appendix Table~\ref{tab:leave_one_out_breakdown}.


\begin{table}[t!]
    \centering
    \caption{\textbf{Task-ID conditioning improves in-distribution specialisation but hinders generalisation.} Results for compositional and unseen transformations, reported as mean $\pm$ s.d. over 5 seeds.}
    \label{tab:generalisation}
    \begin{tabular}{llll}
        \toprule
        & \multicolumn{1}{c}{\textbf{In-Distribution}}
        & \multicolumn{2}{c}{\textbf{Generalisation}} \\
        \cmidrule(lr){2-2} \cmidrule(lr){3-4}
        & Exact Match $\uparrow$
        & Token Accuracy $\uparrow$
        & Exact Match $\uparrow$ \\
        \midrule

        \rowcolor{sectionfill}
\multicolumn{4}{l}{\emph{Compositional}} \\
        Task ID
        & $\mathbf{93.0 \pm 2.4}$
        & $67.7 \pm 1.3$
        & $0.05 \pm 0.10$ \\

        w/o Task ID
        & $57.4 \pm 5.1$ {\color{purple}(\textbf{-35.6})}
        & $\mathbf{75.5 \pm 1.1}$ {\color{linkteal}(\textbf{+7.8})}
        & $\mathbf{1.00 \pm 0.71}$ {\color{linkteal}(\textbf{+0.95})} \\

        \midrule 
        \rowcolor{sectionfill}
\multicolumn{4}{l}{\emph{Unseen}} \\
        Task ID
        & $\mathbf{97.0 \pm 0.5}$
        & $68.5 \pm 4.8$
        & $0.29 \pm 0.57$ \\

        w/o Task ID
        & $64.7 \pm 3.2$ {\color{purple}(\textbf{-32.3})}
        & $\mathbf{81.9 \pm 0.4}$ {\color{linkteal}(\textbf{+13.4})}
        & $\mathbf{9.14 \pm 1.14}$ {\color{linkteal}(\textbf{+8.85})} \\

        \bottomrule
    \end{tabular}
    \vspace{-1em}
\end{table}

\section{Discussion}

\paragraph{From Task Identification to Model Construction.}
Our results suggest that few-shot function induction need not remain within a general-purpose model: a function inferred from demonstrations can instead be instantiated as a compact standalone specialist. Task-ID conditioning improves specialisation to known transformations but hinders generalisation beyond them, suggesting that functions can be inferred directly from demonstrations without explicit task identity. This extends the idea of tool use from selecting predefined tools to generating a specialised executable that can be reused independently. More broadly, a pretrained model could act as a model generator, using demonstrations to construct task-specific specialists on the fly.

\vspace{-0.5em}
\paragraph{Towards Local and Decentralised Execution.}
Generated weights could provide a form of task-specific pretraining, giving a compact model a useful initialisation from only a few demonstrations. Rather than being the final model, this specialist could then be further fine-tuned or personalised using local data before being executed independently. This could reduce reliance on remote inference and allow task- or user-specific data to remain local, with potential benefits for privacy and security. More broadly, a large pretrained model could generate initial specialists that are subsequently refined and deployed on resource-constrained hardware. Our experiments do not yet establish these benefits in latency, memory, compute, privacy, or security, nor how the approach scales beyond tiny target models.

\section{Conclusion}

We study whether few-shot function induction can be realised by generating task-specific model parameters directly from demonstrations. We show that individual ARC-1D transformations require little capacity but incur a substantial cost when represented in shared parameters (Section~\ref{sec:target_capacity}); that a hypernetwork can generate compact specialists whose parameters form a structured weight space (Section~\ref{sec:hypernetwork_generation}); and that these specialists can be reused across new instances and partially generalise to novel compositions and transformations not seen during training (Section~\ref{sec:generalisation}). In both generalisation settings, removing explicit task identity improves transfer beyond the training transformations. Together, these results provide a proof of concept that few-shot task context can be compiled on-the-fly into compact, executable model parameters, separating task inference from execution and enabling reusable specialists that generalise beyond known functions.

\paragraph{Future Directions.}
This proof of concept suggests two directions. First, extending weight generation beyond ARC-1D to richer reasoning settings such as ARC-AGI, including small looped or recursive specialists generated directly from demonstrations without explicit task identifiers. Second, scaling weight generation to larger models through structured or constrained parameter spaces, such as adapters or LoRA updates. Generated parameters could then provide a form of task-specific initialisation, transferring capabilities from a large pretrained model to a compact specialist that can be further refined using downstream or local data. This could enable specialists to be generated on demand, personalised, and executed independently.

\section*{Limitations}

Our evaluation is deliberately controlled: ARC-1D contains a limited set of transformations, so it remains unclear whether weight generation scales to richer reasoning settings such as ARC-AGI or language modelling. Although specialists are generated from only a few demonstrations at inference time, the hypernetwork is trained with substantial data augmentation, so our results demonstrate few-shot task inference rather than learning the weight generator from few examples. We also generate complete parameters for small target networks; scaling to larger models may require structured or constrained parameter spaces or parameter-efficient updates such as adapters or LoRAs. Finally, while generated specialists can be reused across inputs, we do not measure end-to-end efficiency gains in latency, memory, or compute relative to alternative inference-time adaptation methods.

\section*{Broader impacts}
Generating compact specialised models on-the-fly could enable cheaper inference and local deployment, reducing reliance on large-scale GPU infrastructure and remote API services. This may improve accessibility and privacy while reducing computational and energy costs. However, we do not measure these efficiency benefits directly, and establishing them remains future work.

\newpage

{\small \bibliographystyle{plainnat} 
\bibliography{references} 

\appendix

\newpage 

\section{Training Details}
\label{sec:training_details}

\paragraph{Training.} All experiments are implemented in PyTorch Lightning and trained on a single NVIDIA A40 GPU for 8,000 steps, using cosine learning-rate decay with linear warmup. We use Muon~\citep{jordan2024muon,liu2025muon} for hidden 2-D linear weights and AdamW~\citep{loshchilov2018decoupled} for embeddings, input/output layers, biases, normalisation, and convolution parameters, with optimisation hyperparameters given in Table~\ref{tab:training_hyperparameters}. Muon uses momentum 0.95 with Nesterov momentum and Newton--Schulz orthogonalisation. Directly trained models use batches of 256 input--output pairs in FP32 with TF32 matrix multiplications, while hypernetworks use batches of 512 episodes with BF16 mixed precision. All Transformers use dropout 0.1, and validation is performed once per epoch. We train five seeds per configuration and report mean $\pm$ standard error. Models selected by validation use the checkpoint with the lowest validation loss.

\begin{table}[h!]
    \centering
    \caption{Optimisation hyperparameters for directly trained models and hypernetworks.}
    \label{tab:training_hyperparameters}
    \begin{tabular}{lcc}
        \toprule
        & Direct models & Hypernetworks \\
        \midrule
        Muon learning rate & 0.005 & 0.005 \\
        Muon momentum & 0.95 & 0.95 \\
        AdamW learning rate & 0.0005 & 0.001 \\
        Weight decay & 0.01 & 0.01 \\
        Warmup steps & 200 & 800 \\
        Cosine minimum LR & 0 & $10^{-4}$ \\
        Gradient clipping & None & 10.0 \\
        \bottomrule
    \end{tabular}
\end{table}

\paragraph{Model comparison.}

We evaluate the three-layer RoPE~\citep{su_2024_rope} Canon
Transformer~\citep{allen-zhu2025physics,vaswani2017} at hidden dimensions
$4$, $6$, and $10$, corresponding to approximately 1.4K, 2.4K, and 5.4K
parameters. Individual models are trained separately for each transformation
using five seeds per model size. Hypernetworks are trained jointly across all
14 transformations to generate the same target architectures, either from
the support examples alone or with a frozen task embedding, with results
averaged over five seeds. All experiments use the augmented ARC-1D dataset
described in Section~\ref{sec:data}. Canon layers use SiLU activations, while
MLP blocks and the hypernetwork decoder use GELU.

\begin{table}[h]
    \centering
    \caption{\textbf{Model configurations.} Direct models are the three-layer RoPE Canon Transformer at four widths. The hypernetwork generates every weight of the target except the token embedder, which it shares with its encoder. Only the dimension-4 hypernetwork is trained; grey values marked with $\dagger$ show the corresponding parameter counts for wider targets, where only the decoder output layer grows.}
    \label{tab:architectures}
    \begin{tabular}{lcccc}
        \toprule
        Hidden dimension & 4 & 6 & 10 & 14 \\
        \midrule
        \multicolumn{5}{l}{\textbf{Direct model}} \\
        Layers / heads            & 3 / 1 & 3 / 1 & 3 / 1 & 3 / 1 \\
        Token embedding dim.      & 10 & 10 & 10 & 10 \\
        Canon kernel              & 5 & 5 & 5 & 5 \\
        Parameters                & 1,398 & 2,444 & 5,400 & 9,508 \\
        \midrule
        \multicolumn{5}{l}{\textbf{Hypernetwork}} \\
        Token embedding dim.      & 4 & 4 & 4 & 4 \\
        Encoder dim. / layers     & 4 / 1 & 4 / 1 & 4 / 1 & 4 / 1 \\
        Decoder bottleneck ($d'$) & 8 & 8 & 8 & 8 \\
        Generated target weights  & 1,204 & \textcolor{gray}{2,238$^\dagger$} & \textcolor{gray}{5,170$^\dagger$} & \textcolor{gray}{9,254$^\dagger$} \\
        Trainable parameters      & 10,156 & \textcolor{gray}{18,428$^\dagger$} & \textcolor{gray}{41,884$^\dagger$} & \textcolor{gray}{74,556$^\dagger$} \\
        \bottomrule
    \end{tabular}
\end{table}

\newpage 

\paragraph{Compute.} All runs use a single NVIDIA A40 GPU (48\,GB). Table~\ref{tab:compute} reports the number of runs, wall-clock time per run including data setup and evaluation, and total GPU-hours for each experiment. Directly trained models take approximately 3--9 minutes per run and hypernetworks approximately 24 minutes for 8,000 training steps, although the data-efficiency experiments include substantially longer runs. In total, the experiments reported in this paper comprise 2,000 runs and approximately 460 GPU-hours.

\begin{table}[h]
    \centering
    \caption{\textbf{Compute per experiment.} Wall-clock time per run and total compute on a single NVIDIA A40 GPU unless otherwise indicated. Times include data setup and evaluation.}
    \label{tab:compute}
    \begin{tabular}{llrrr}
        \toprule
        & & Runs & Min./run & GPU-h \\
        \midrule
        1 & Capacity and ablations & 960 & 3--9 & 64 \\
        2 & Hypernetwork, in distribution & 10 & 24 & 4 \\
        3 & Weight reuse (evaluation only) & 10 & 0.4 & $<1$ \\
        4 & Compositional generalisation (CPU evaluation) & 10 & 0.7 & 0 \\
        5 & Leave-one-out & 140 & 23 & 54 \\
        6 & Data efficiency & 870 & 7--109 & 340 \\
        \midrule
        & \textbf{Total} & \textbf{2,000} & & \textbf{461} \\
        \bottomrule
    \end{tabular}
\end{table}

\newpage
\section{Architecture \& Optimiser Ablations}
\label{sec:ablations}

\paragraph{Task-Identity Conditioning.} We ask where task identity should enter the hypernetwork and whether its representation should be learned. Our main model adds a frozen random projection of the one-hot task ID to the pooled task representation $z$. We compare this latent conditioning with adding the task embedding to every support token before the encoder, using either frozen random or learned embeddings, alongside a model without task identity. Training follows Section~\ref{sec:hypernetwork_generation}, while generalisation follows Section~\ref{sec:generalisation}: held-out categories receive a zero task vector, and unseen compositions receive the sum of their constituent task vectors.

\begin{table}[h]
    \centering
    \caption{\textbf{Task-ID conditioning improves in-distribution specialisation but hinders generalisation.} Results across task-ID placements and embedding types, reported as mean $\pm$ s.d.\ over 5 seeds. Parentheses show the change relative to no task-ID conditioning.}
    \label{tab:task_id}
    \begin{tabular}{lllll}
        \toprule
        & \multicolumn{1}{c}{\textbf{In-Distribution}}
        & \multicolumn{3}{c}{\textbf{Generalisation}} \\
        \cmidrule(lr){2-2} \cmidrule(lr){3-5}
        & Exact Match $\uparrow$
        & \multicolumn{2}{c}{\textbf{Unseen}}
        & \multicolumn{1}{c}{\textbf{Compositional}} \\
        \cmidrule(lr){3-4} \cmidrule(lr){5-5}
        & 
        & Token Accuracy $\uparrow$
        & Exact Match $\uparrow$
        & Token Accuracy $\uparrow$ \\
        \midrule

        None
        & $57.4 \pm 5.7$
        & $\mathbf{81.6 \pm 0.9}$
        & $\mathbf{10.2 \pm 3.6}$
        & $\mathbf{75.5 \pm 1.2}$ \\

        \midrule
        \rowcolor{sectionfill}
        \multicolumn{5}{l}{\emph{Latent conditioning}} \\

        Frozen
        & $93.0 \pm 2.6$ {\color{linkteal}(\textbf{+35.6})}
        & $79.1 \pm 1.6$ {\color{purple}(\textbf{-2.5})}
        & $0.6 \pm 0.5$ {\color{purple}(\textbf{-9.6})}
        & $63.2 \pm 8.3$ {\color{purple}(\textbf{-12.3})} \\

        Learned
        & $\mathbf{96.5 \pm 0.8}$ {\color{linkteal}(\textbf{+39.1})}
        & $77.3 \pm 2.9$ {\color{purple}(\textbf{-4.3})}
        & $1.1 \pm 1.9$ {\color{purple}(\textbf{-9.1})}
        & $66.4 \pm 4.8$ {\color{purple}(\textbf{-9.1})} \\

        \midrule
        \rowcolor{sectionfill}
        \multicolumn{5}{l}{\emph{Input conditioning}} \\

        Frozen
        & $95.1 \pm 2.2$ {\color{linkteal}(\textbf{+37.7})}
        & $74.1 \pm 4.2$ {\color{purple}(\textbf{-7.5})}
        & $0.0 \pm 0.0$ {\color{purple}(\textbf{-10.2})}
        & $61.0 \pm 3.0$ {\color{purple}(\textbf{-14.5})} \\

        Learned
        & $96.2 \pm 1.3$ {\color{linkteal}(\textbf{+38.8})}
        & $73.1 \pm 6.2$ {\color{purple}(\textbf{-8.5})}
        & $0.8 \pm 0.8$ {\color{purple}(\textbf{-9.4})}
        & $68.8 \pm 2.9$ {\color{purple}(\textbf{-6.7})} \\

        \bottomrule
    \end{tabular}
\end{table}

All task-ID variants substantially improve in-distribution performance, but generalise less well than the model without task identity (Table~\ref{tab:task_id}). Learning the task embedding provides little additional benefit over a frozen random embedding. Placement has a clearer effect: latent conditioning generally retains better generalisation than conditioning the encoder input, while preserving similarly strong in-distribution performance. The representations in Figure~\ref{fig:task_id_clusters} are consistent with this difference: latent conditioning produces a clearer task-aligned organisation, whereas input conditioning yields a coarser structure. Overall, explicit task identity improves specialisation to known transformations, but neither its placement nor learning the embedding recovers the generalisation obtained without task-ID conditioning.

\begin{figure}[h!]
    \centering
    \includegraphics[width=\linewidth]{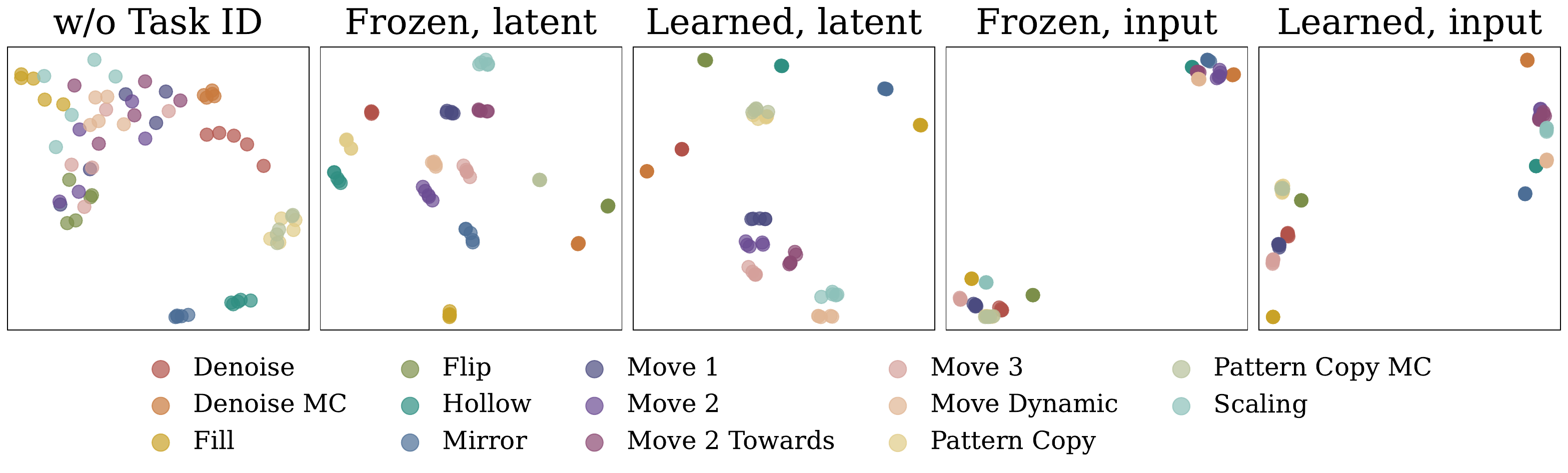}
    \caption{\textbf{Task representations depend on task-ID placement.} t-SNE of the task representation $z$ for validation examples from each transformation (seed 1; axes are arbitrary). Latent conditioning produces clearer task-aligned structure, while input conditioning forms coarser groups.}
    \label{fig:task_id_clusters}
\end{figure}
\vspace{-1em}

\paragraph{Canon ablation.} We ablate the Canon layers across all Section \ref{sec:target_capacity} configurations, covering individual and joint models with and without task identity at all four model sizes, using five seeds per configuration. Canon improves test exact-match accuracy across all settings, with the largest gains at low capacity (Figure~\ref{fig:capacity_cliff_canons}). These results indicate that the Canon layers are particularly important when capacity is limited or shared across transformations.

\newpage

\begin{figure}[h!]
    \centering
    \includegraphics[width=0.48\linewidth]{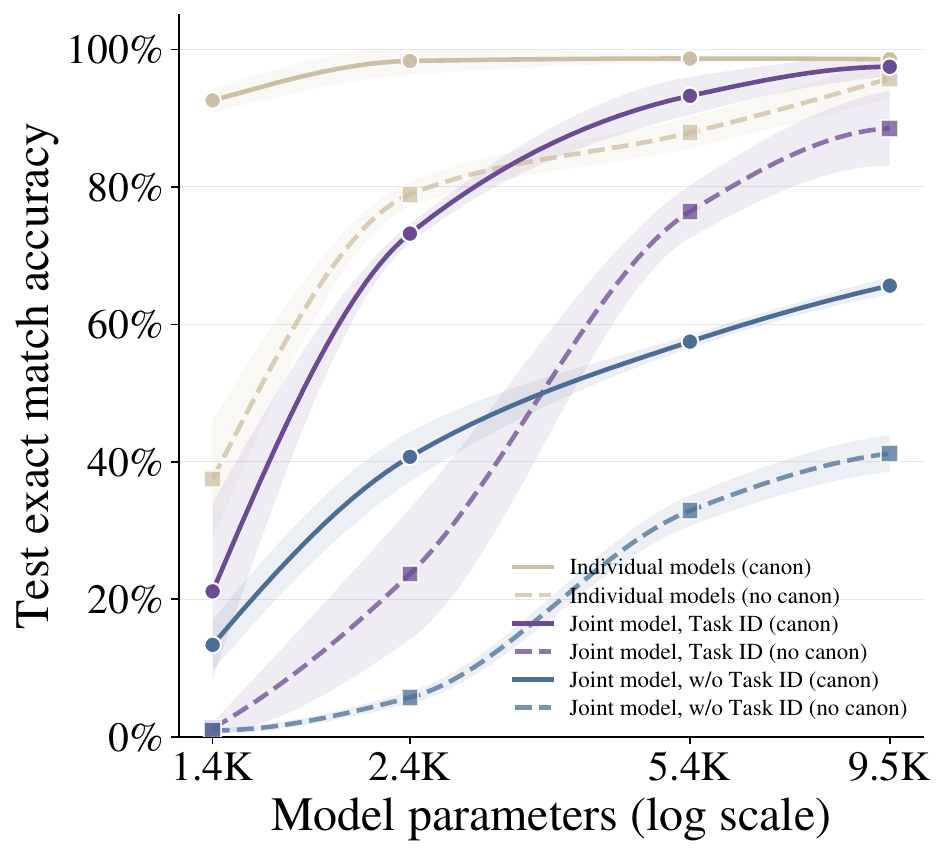}
    \caption{\textbf{Canon layers improve performance, particularly at low capacity.} Test exact-match accuracy with and without Canon layers for individual and joint models across model sizes. The gain diminishes for larger individual models but remains substantial for jointly trained models. Bands show $\pm1$ s.d. across 5 seeds.}
    \label{fig:capacity_cliff_canons}
\end{figure}
\vspace{-1em}

\paragraph{Optimiser ablation.} We replace Muon with AdamW in every configuration of Section~\ref{sec:target_capacity}, using five seeds per configuration. Muon uses a learning rate of 0.005 with Nesterov momentum 0.95 and an auxiliary AdamW group (learning rate 0.0005, $\beta=(0.9,0.95)$) for embeddings, the output layer, normalisation, and Canon parameters. AdamW instead optimises all parameters with a learning rate of 0.001 and $\beta=(0.9,0.999)$. Both use weight decay 0.01 and the same warmup and cosine schedule. Muon matches or exceeds AdamW in every setting (Figure~\ref{fig:capacity_cliff_muon}). Note: the AdamW learning rate was not independently tuned, this ablation does not isolate optimiser choice from hyperparameter selection. We look at this a bit more closely for the hypernetwork architectures.

\begin{figure}[h!]
    \centering
    \includegraphics[width=0.48\linewidth]{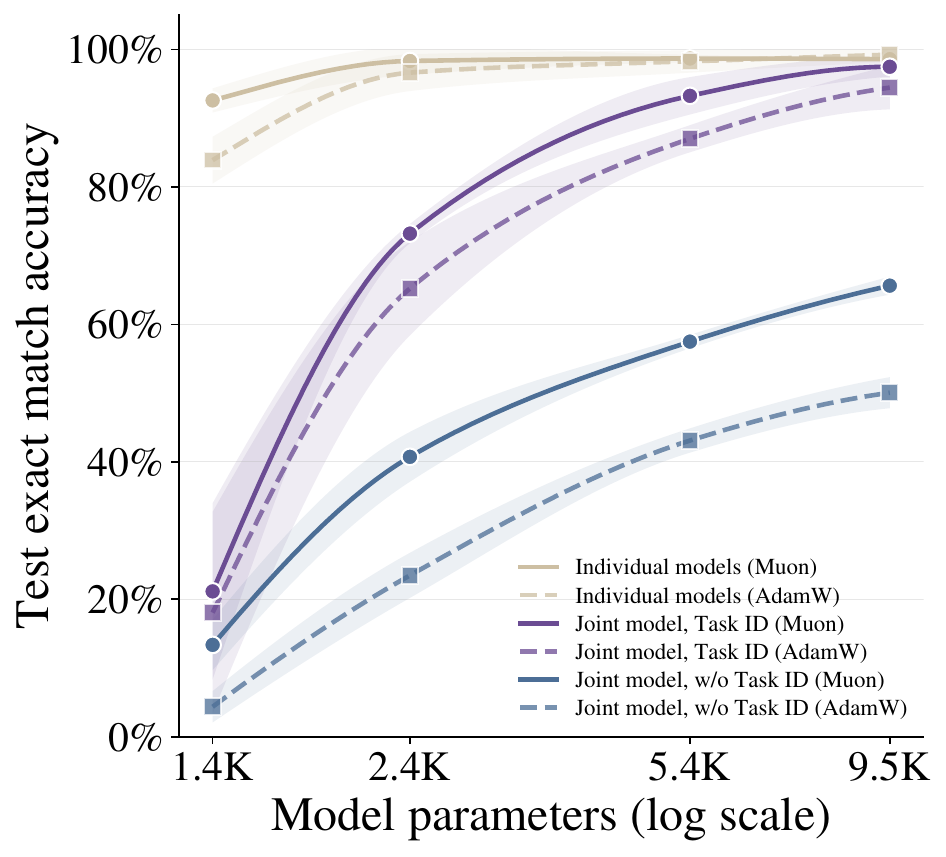}
    \caption{\textbf{Muon improves marginally over AdamW} Test exact-match accuracy with Muon (solid) and AdamW (dashed) across model sizes. Bands show $\pm1$ s.d.\ across 5 seeds.}
    \label{fig:capacity_cliff_muon}
\end{figure}

\paragraph{Hyperparameter search for hypernetworks.} We tried tuning both optimisers on the hypernetwork without task-identity conditioning, using three seeds per configuration. For Muon, we search the Muon learning rate ($\{0.002, 0.005, 0.01, 0.02, 0.04\}$), auxiliary AdamW learning rate ($\{10^{-4}, 3{\times}10^{-4}, 5{\times}10^{-4}, 10^{-3}\}$), and weight decay ($\{0, 0.01, 0.1\}$), giving 60 configurations. We fix momentum to 0.95 with Nesterov momentum, use five Newton--Schulz steps, and set the auxiliary AdamW betas to $(0.9,0.95)$. For AdamW, we search the learning rate ($\{10^{-4}, 3{\times}10^{-4}, 10^{-3}, 3{\times}10^{-3}, 10^{-2}\}$) and weight decay ($\{0, 0.01, 0.1\}$), giving 15 configurations, with betas $(0.9,0.95)$. All configurations are trained for 8,000 steps with 800 warmup steps and cosine decay to zero, and are selected by validation loss. Muon is relatively insensitive to the searched hyperparameters: its best configurations achieve the highest test exact-match accuracy, reaching $66.7 \pm 11.0\%$, but with high variation across seeds. AdamW achieves a lower best validation loss and a test exact-match accuracy of $60.0 \pm 0.9\%$.

\section{Task Performance Individual vs Joint}
\label{sec:per_task_plots}

\begin{figure}[h!]
    \centering
    \includegraphics[width=0.7\linewidth]{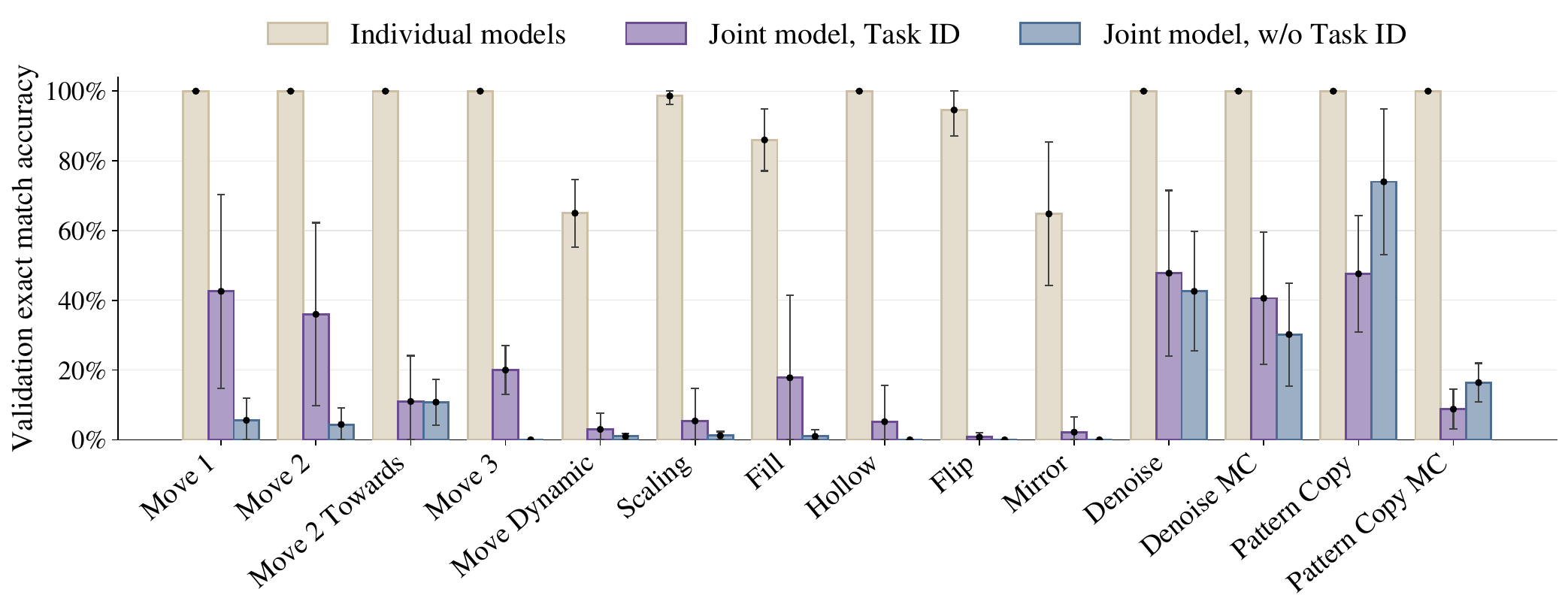}
    \caption{\textbf{The joint-training penalty is concentrated in a few
    transformations.} Per-transformation exact-match validation accuracy at
    1.4K parameters. Error bars show $\pm1$ s.d. across 5 seeds.}
    \label{fig:per_task_dim4}
\end{figure}

\begin{figure}[h!]
    \centering
    \includegraphics[width=0.7\linewidth]{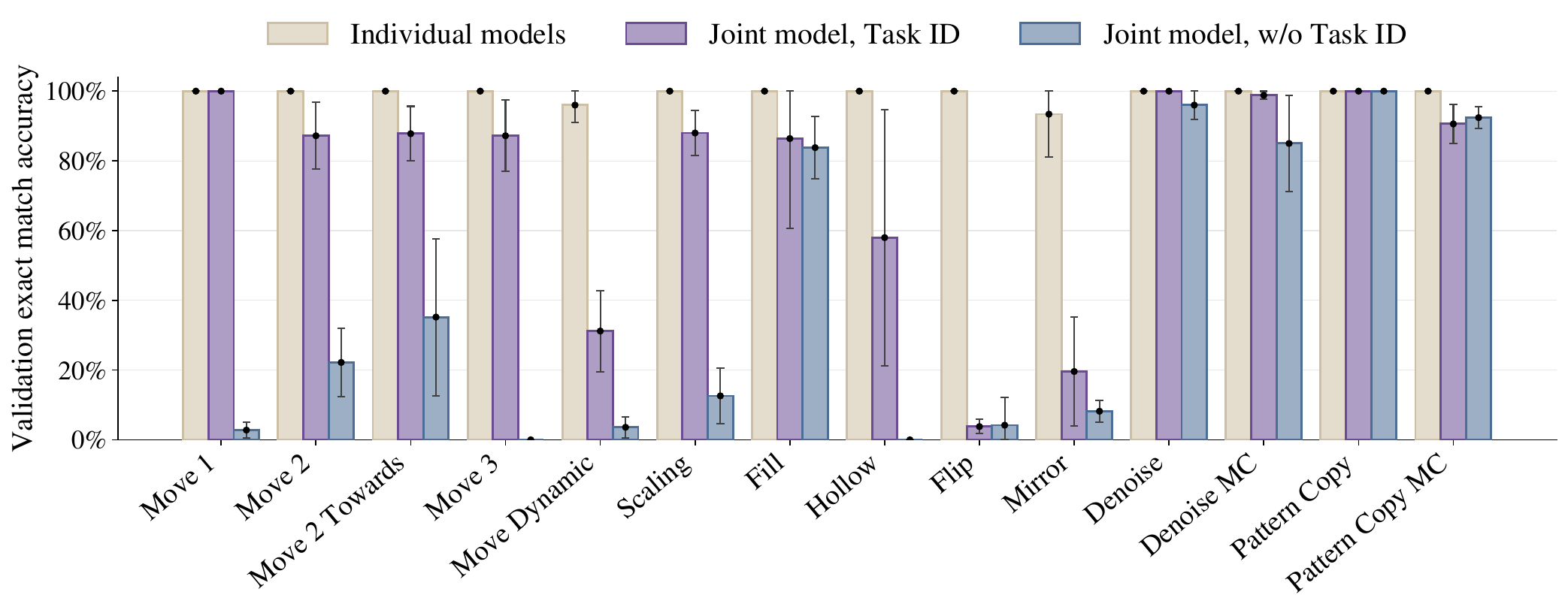}
    \caption{\textbf{The joint-training penalty is concentrated in a few
    transformations.} Per-transformation exact-match validation accuracy at
    2.4K parameters. Error bars show $\pm1$ s.d. across 5 seeds.}
    \label{fig:per_task_dim6}
\end{figure}

\begin{figure}[h!]
    \centering
    \includegraphics[width=0.7\linewidth]{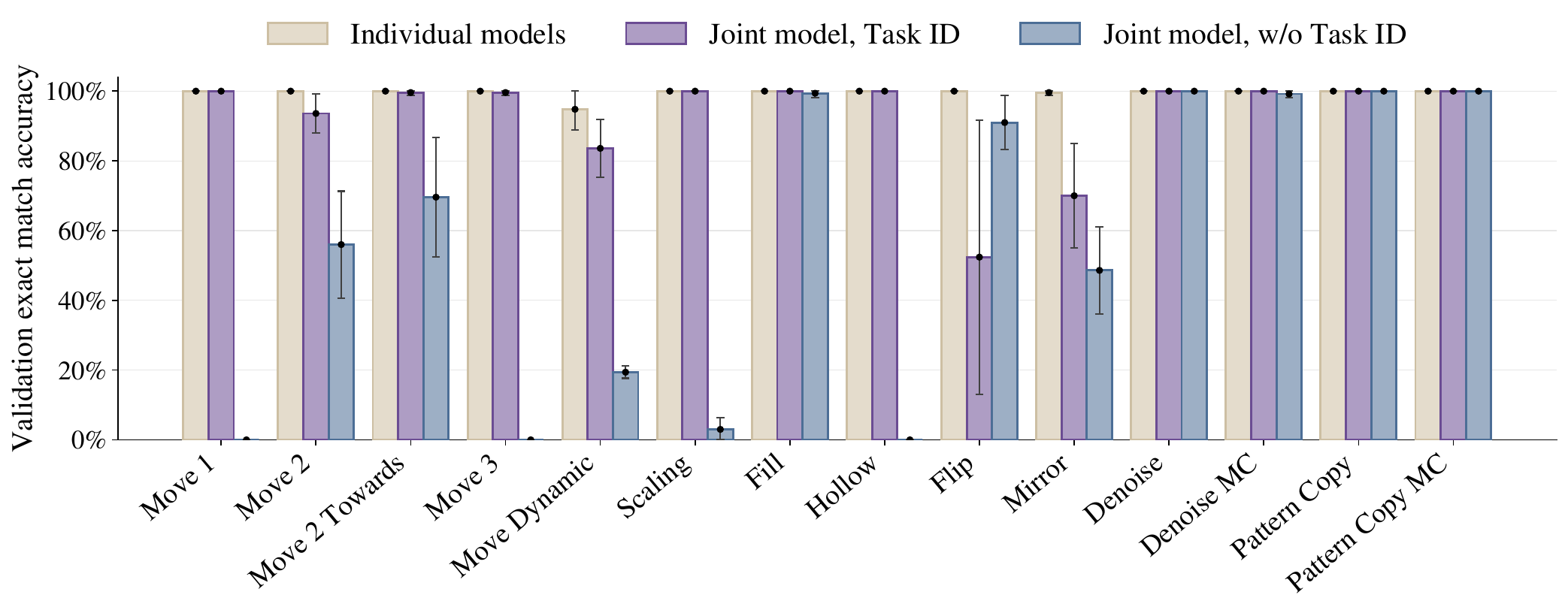}
    \caption{\textbf{The joint-training penalty is concentrated in a few
    transformations.} Per-transformation exact-match validation accuracy at
    5.4K parameters. Error bars show $\pm1$ s.d. across 5 seeds.}
    \label{fig:per_task_dim10}
\end{figure}

\begin{figure}[h!]
    \centering
    \includegraphics[width=0.7\linewidth]{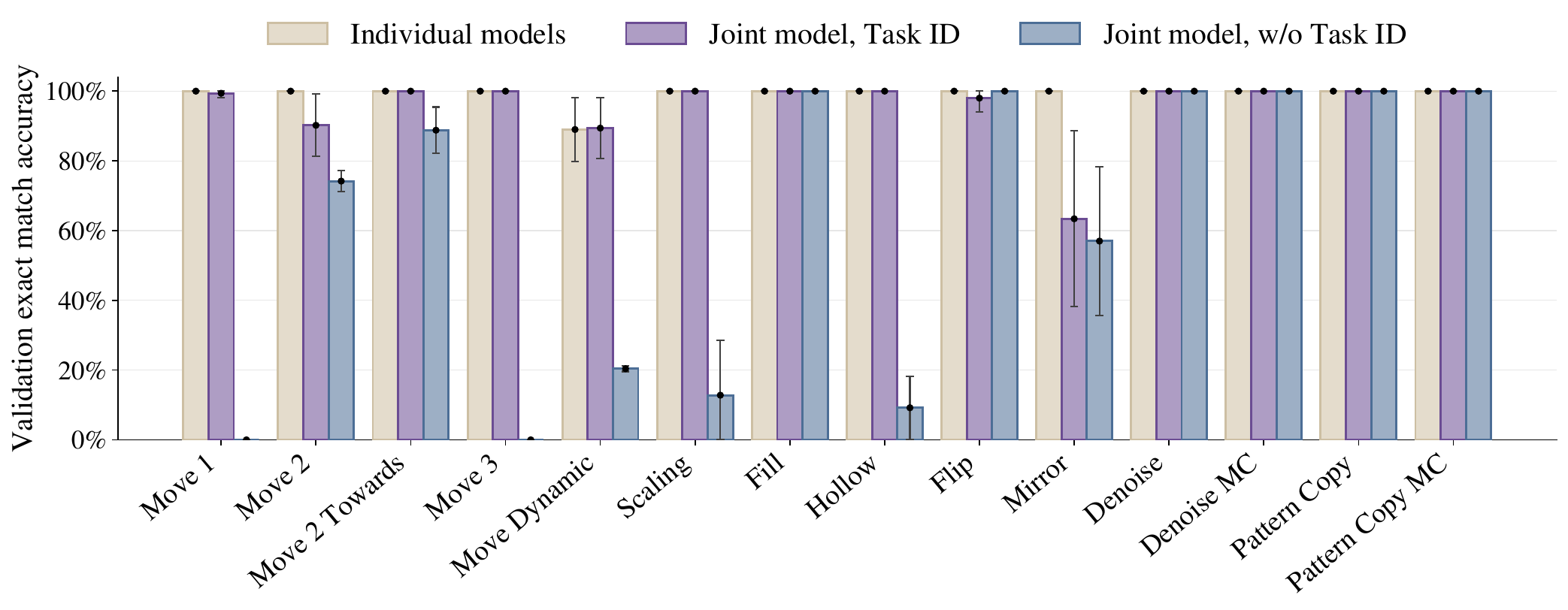}
    \caption{\textbf{The joint-training penalty is concentrated in a few
    transformations.} Per-transformation exact-match validation accuracy at
    9.5K parameters. Error bars show $\pm1$ s.d. across 5 seeds.}
    \label{fig:per_task_dim14}
\end{figure}

\newpage

\section{Data Efficiency}
\label{sec:data_efficiency}

We ask whether sharing across transformations improves data efficiency. ARC-1D provides 50 episodes per transformation, of which 40 are used for training, and we increase the training set through augmentation. We compare individual models, hypernetworks, and joint models, with and without task-ID conditioning (Figure~\ref{fig:data_efficiency}). Individual models and task-ID-conditioned hypernetworks follow similar data-efficiency curves, showing no material benefit from hypernetwork weight sharing at this scale. The task-ID-conditioned joint model performs better with less data, but stores all transformations directly in its 10K parameters rather than generating decoupled 1.4K-parameter specialists. Without task IDs, the joint model and hypernetwork perform similarly across training-set sizes. Across models, augmentation provides diminishing gains beyond roughly 100--200 training episodes per transformation. Similarly, \citet{schurholt2025hrm} find that small HRMs on ARC-AGI-1 benefit from augmentation but require substantially fewer than 1K augmentations. We retain 1K augmentations in our main experiments for consistency.

\begin{figure}[h!]
    \centering
    \includegraphics[width=0.48\linewidth]{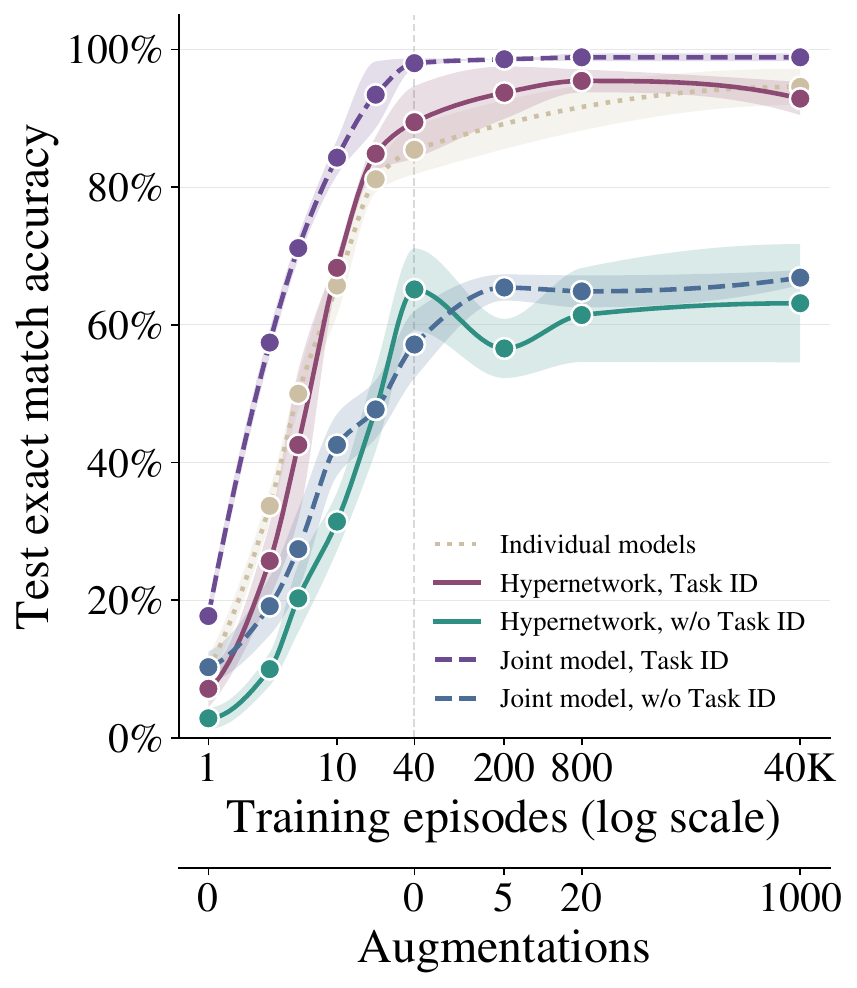}
\caption{\textbf{Data efficiency across models.} Individual models and task-ID-conditioned hypernetworks require similar amounts of training data, while the larger task-ID-conditioned joint model learns with fewer episodes. Without task IDs, joint models and hypernetworks follow similar curves. Bands show $\pm1$ s.d. across 5 seeds.}
    \label{fig:data_efficiency}
\end{figure}

\newpage

\section{Generalisation Details}
\label{sec:comp_gen_details}

\begin{table}[h]
    \centering
    \caption{\textbf{Performance on held-out compositions.}
    Each composition combines two transformations observed independently during
    training, while the composition itself is held out.
    $B \circ A$ denotes applying $A$ first, followed by $B$.
    Token accuracy is macro-averaged over the 10 compositions and 5 seeds;
    exact match is pooled over all 2,000 held-out instances.}
    \label{tab:compositional_breakdown}
    \begin{tabular}{lllll}
        \toprule
        & \multicolumn{2}{c}{\textbf{Task ID}}
        & \multicolumn{2}{c}{\textbf{w/o Task ID}} \\
        \cmidrule(lr){2-3} \cmidrule(lr){4-5}
        \textbf{Composition}
        & \textbf{Token Acc.}
        & \textbf{EM}
        & \textbf{Token Acc.}
        & \textbf{EM} \\
        \midrule

        $\text{Mirror} \circ \text{Denoise MC}$
            & 47.7 & 0.0 & \textbf{85.6}\ {\color{linkteal}(\textbf{+37.9})} & \textbf{8.0} \\

        $\text{Copy} \circ \text{Denoise MC}$
            & \textbf{69.6} & 0.0 & 67.3\ {\color{purple}(\textbf{-2.3})} & 0.0 \\

        $\text{Denoise 1C} \circ \text{Denoise MC}$
            & \textbf{76.7} & 0.0 & 68.3\ {\color{purple}(\textbf{-8.3})} & \textbf{0.5} \\

        $\text{Mirror} \circ \text{Fill}$
            & 61.3 & 0.0 & \textbf{73.1}\ {\color{linkteal}(\textbf{+11.7})} & \textbf{1.5} \\

        $\text{Shift 3} \circ \text{Denoise 1C}$
            & 75.5 & \textbf{0.5} & \textbf{77.0}\ {\color{linkteal}(\textbf{+1.5})} & 0.0 \\

        $\text{Shift 3} \circ \text{Hollow}$
            & 78.2 & 0.0 & \textbf{85.2}\ {\color{linkteal}(\textbf{+7.0})} & 0.0 \\

        $\text{Hollow} \circ \text{Move Dynamic}$
            & 79.1 & 0.0 & \textbf{82.4}\ {\color{linkteal}(\textbf{+3.2})} & 0.0 \\

        $\text{Shift 3} \circ \text{Fill}$
            & 68.8 & 0.0 & \textbf{73.9}\ {\color{linkteal}(\textbf{+5.1})} & 0.0 \\

        $\text{Move Dynamic} \circ \text{Fill}$
            & 62.7 & 0.0 & \textbf{71.6}\ {\color{linkteal}(\textbf{+9.0})} & 0.0 \\

        $\text{Copy} \circ \text{Shift 3}$
            & 57.2 & 0.0 & \textbf{70.3}\ {\color{linkteal}(\textbf{+13.1})} & 0.0 \\

        \midrule
        \textbf{Overall}
            & 67.7 & 0.05
            & \textbf{75.5\ {\color{linkteal}(+7.8)}} & \textbf{1.00} \\
        \bottomrule
    \end{tabular}
\end{table}

\begin{table}[h]
    \centering
    \caption{\textbf{Performance on held-out task categories.}
    Each row holds out one of the 14 base transformations entirely during training and
    evaluates it zero-shot. Token accuracy and exact match are each the mean across 5 seeds
    for that category's own held-out score.}
    \label{tab:leave_one_out_breakdown}
    \begin{tabular}{lllll}
        \toprule
        & \multicolumn{2}{c}{\textbf{Task ID}}
        & \multicolumn{2}{c}{\textbf{w/o Task ID}} \\
        \cmidrule(lr){2-3} \cmidrule(lr){4-5}
        \textbf{Category}
        & \textbf{Token Acc.}
        & \textbf{EM}
        & \textbf{Token Acc.}
        & \textbf{EM} \\
        \midrule

        Denoise Multicolor
            & 47.9 & 0.00 & \textbf{81.1}\ {\color{linkteal}(\textbf{+33.2})} & \textbf{4.00} \\

        Scaling
            & 57.1 & 0.00 & \textbf{86.9}\ {\color{linkteal}(\textbf{+29.8})} & 0.00 \\

        Pattern Copy Multicolor
            & 57.6 & 0.00 & \textbf{84.7}\ {\color{linkteal}(\textbf{+27.2})} & 0.00 \\

        Move 3 Pixels
            & 68.4 & 0.00 & \textbf{91.8}\ {\color{linkteal}(\textbf{+23.4})} & 0.00 \\

        Pattern Copy
            & 77.3 & 0.00 & \textbf{99.5}\ {\color{linkteal}(\textbf{+22.2})} & \textbf{96.00} \\

        Move 2 Pixels Towards
            & 74.9 & 0.00 & \textbf{95.5}\ {\color{linkteal}(\textbf{+20.6})} & \textbf{20.00} \\

        Mirror
            & 59.0 & 0.00 & \textbf{68.1}\ {\color{linkteal}(\textbf{+9.1})} & 0.00 \\

        Move Dynamic
            & 77.0 & 4.00 & \textbf{86.1}\ {\color{linkteal}(\textbf{+9.1})} & \textbf{8.00} \\

        Move 1 Pixel
            & 84.0 & 0.00 & \textbf{92.0}\ {\color{linkteal}(\textbf{+8.0})} & 0.00 \\

        Move 2 Pixels
            & 84.6 & 0.00 & \textbf{91.2}\ {\color{linkteal}(\textbf{+6.5})} & 0.00 \\

        Fill
            & 62.8 & 0.00 & \textbf{69.2}\ {\color{linkteal}(\textbf{+6.4})} & 0.00 \\

        Flip
            & 78.2 & 0.00 & \textbf{83.6}\ {\color{linkteal}(\textbf{+5.5})} & 0.00 \\

        Hollow
            & 66.1 & 0.00 & \textbf{69.9}\ {\color{linkteal}(\textbf{+3.9})} & 0.00 \\

        Denoise
            & \textbf{64.6} & 0.00 & 46.9\ {\color{purple}(\textbf{-17.7})} & 0.00 \\

        \midrule
        \textbf{Overall}
            & 68.5 & 0.29
            & \textbf{81.9\ {\color{linkteal}(+13.4)}} & \textbf{9.14} \\
        \bottomrule
    \end{tabular}
\end{table}

\newpage
\section{Task structured Weight-Space}

\begin{figure}[h!]
    \centering
    \includegraphics[width=\linewidth]{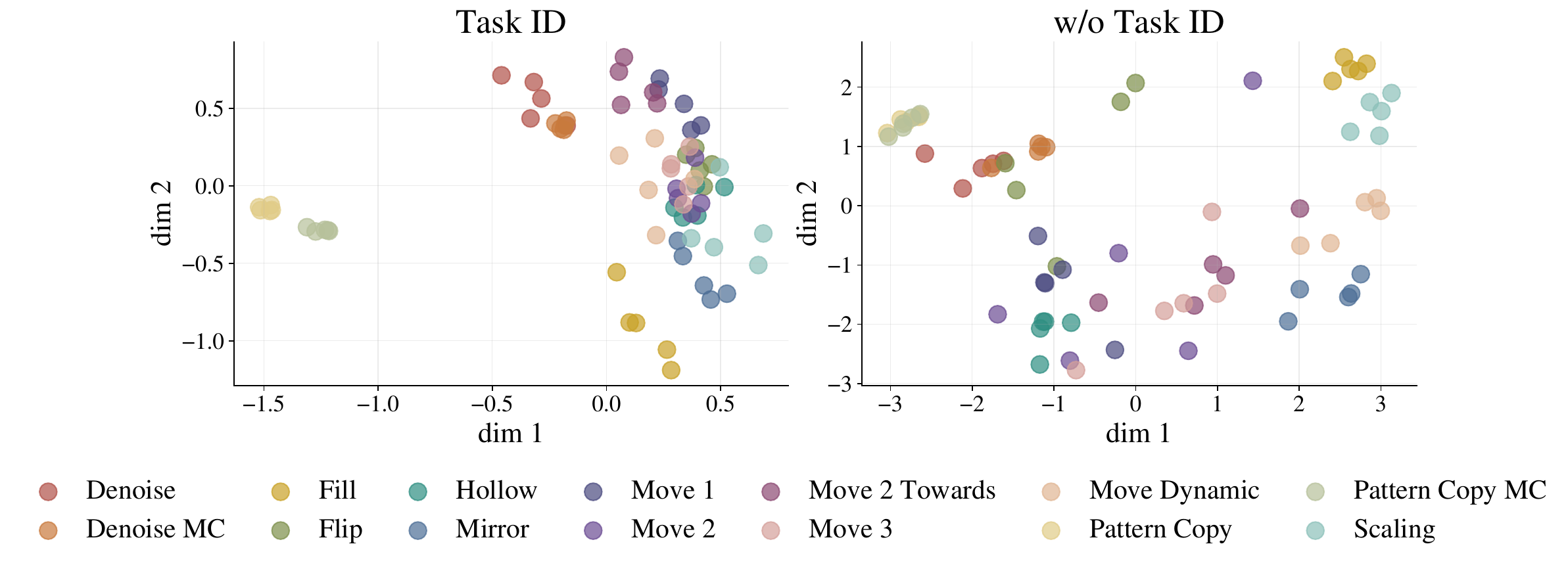}
    \caption{\textbf{Generated parameters form a task-structured weight space.}
PCA projection of generated target-model parameters, coloured by transformation.
\textbf{Left:} With a frozen task embedding, tasks are fully separable (100\% linear probe accuracy).
\textbf{Right:} Without task identity, structure emerges from the support examples alone (60.9\% linear probe accuracy). }
    \label{fig:pca}
\end{figure}

\begin{figure}[h!]
    \centering
    \includegraphics[width=\linewidth]{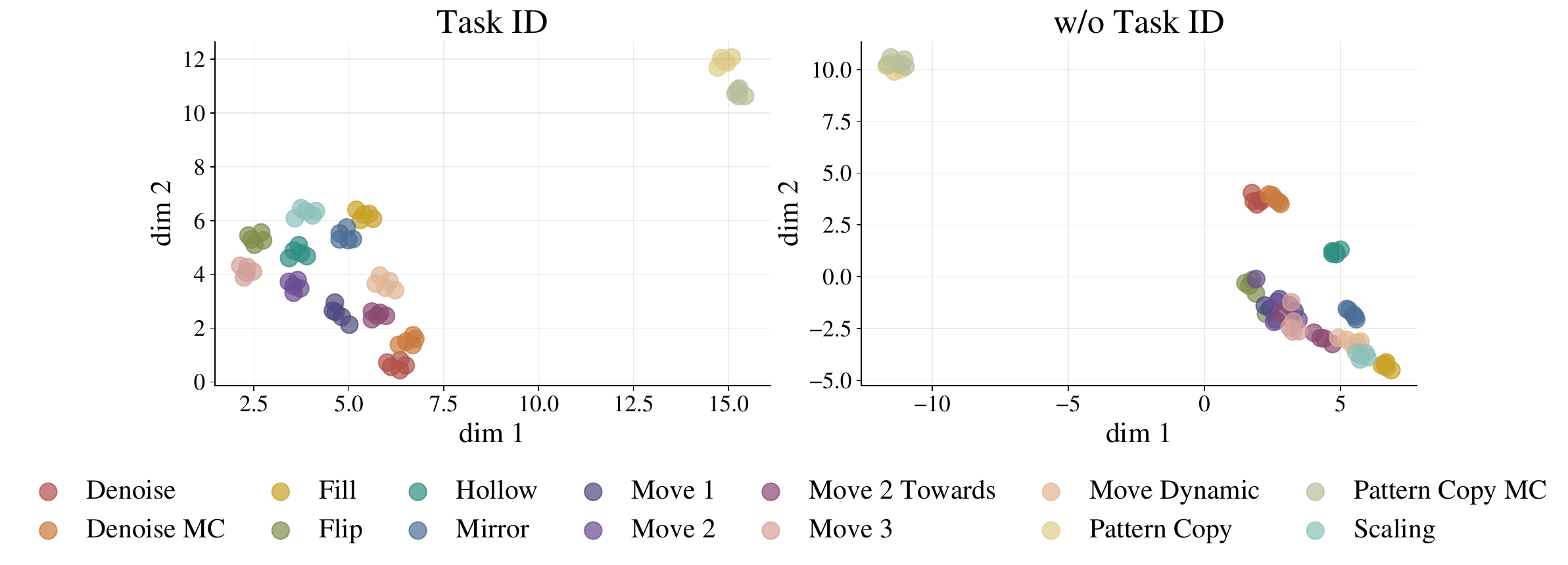}
    \caption{\textbf{Generated parameters form a task-structured weight space.}
UMAP projection of generated target-model parameters, coloured by transformation.
\textbf{Left:} With a frozen task embedding, tasks are fully separable (100\% linear probe accuracy).
\textbf{Right:} Without task identity, structure emerges from the support examples alone (60.9\% linear probe accuracy). }
    \label{fig:umap}
\end{figure}

\newpage 
\section{Task Examples}
\label{sec:task_examples}

\begin{figure}[H]
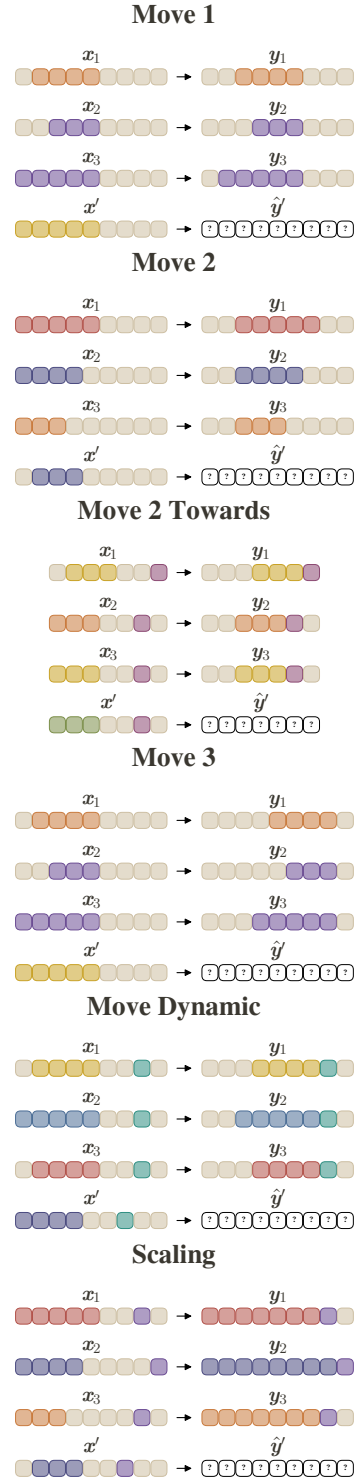

    \centering

    \includegraphics[width=\linewidth]
    {figures/arc1d/tasks/arc_1d_task_1d_move_1p_8.pdf}

    \includegraphics[width=\linewidth]
    {figures/arc1d/tasks/arc_1d_task_1d_move_2p_27.pdf}

    \includegraphics[width=\linewidth]
    {figures/arc1d/tasks/arc_1d_task_1d_move_2p_dp_7.pdf}

    \includegraphics[width=\linewidth]
    {figures/arc1d/tasks/arc_1d_task_1d_move_3p_8.pdf}

    \includegraphics[width=\linewidth]
    {figures/arc1d/tasks/arc_1d_task_1d_move_dp_37.pdf}

    \includegraphics[width=\linewidth]
    {figures/arc1d/tasks/arc_1d_task_1d_scale_dp_27.pdf}

    \caption{\textbf{ARC-1D movement and scaling tasks.}
    Examples of the movement and scaling task families, including one-point
    (1P), two-point (2P), two-point dynamic-position (2P-DP), three-point
    (3P), dynamic-position (DP), and scaling variants.}
    \label{fig:arc1d-movement-scale}
\end{figure}
\newpage

\begin{figure}[H]
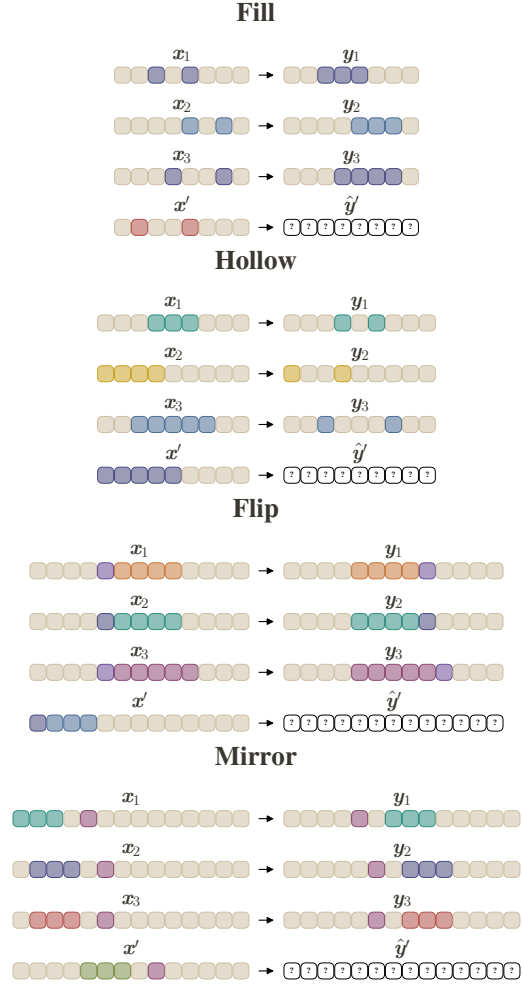

    \centering

    \includegraphics[width=\linewidth]
    {figures/arc1d/tasks/arc_1d_task_1d_fill_25.pdf}

    \includegraphics[width=\linewidth]
    {figures/arc1d/tasks/arc_1d_task_1d_hollow_28.pdf}

    \includegraphics[width=\linewidth]
    {figures/arc1d/tasks/arc_1d_task_1d_flip_46.pdf}

    \includegraphics[width=\linewidth]
    {figures/arc1d/tasks/arc_1d_task_1d_mirror_46.pdf}

    \caption{\textbf{ARC-1D object-transformation tasks.}
    Examples of the fill, hollow, flip, and mirror task families.}
    \label{fig:arc1d-object-transformations}
\end{figure}
\newpage
\begin{figure}[H]
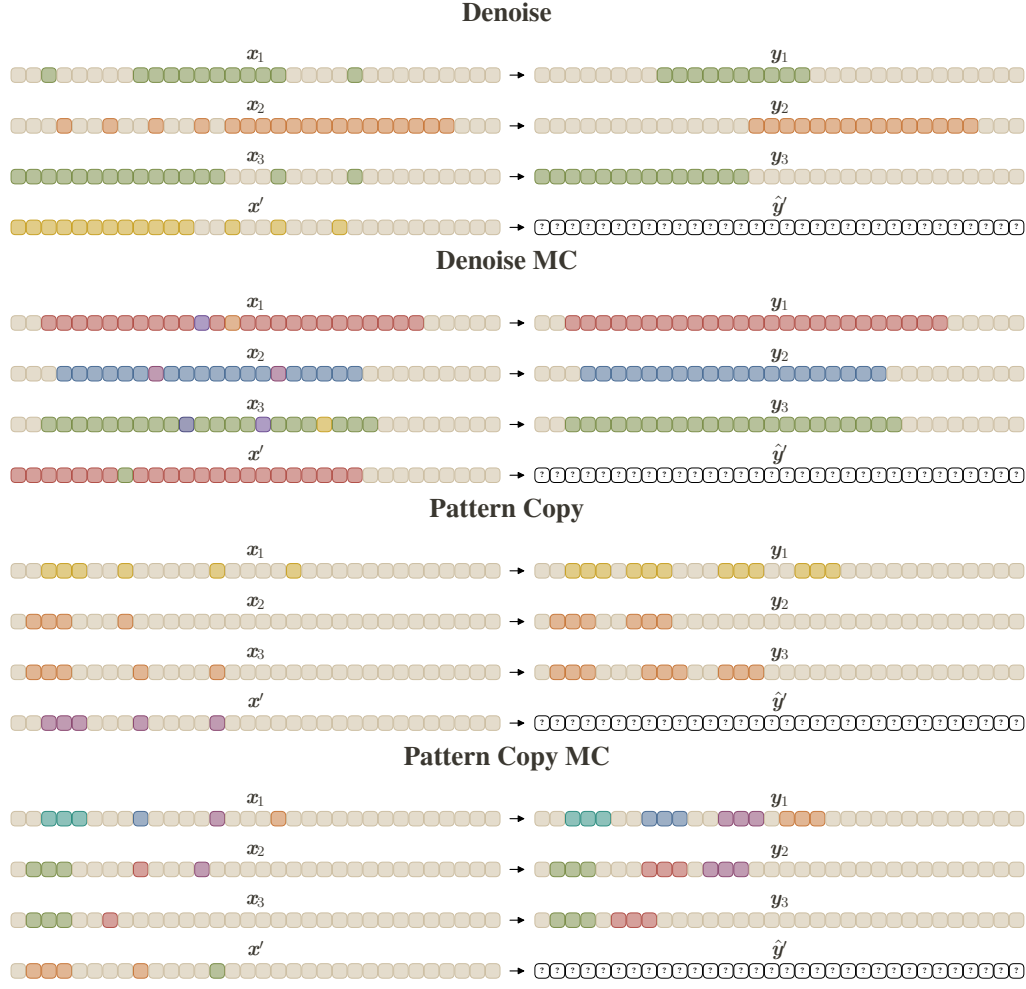

    \centering

    \includegraphics[width=\linewidth]
    {figures/arc1d/tasks/arc_1d_task_1d_denoising_1c_0.pdf}

    \includegraphics[width=\linewidth]
    {figures/arc1d/tasks/arc_1d_task_1d_denoising_mc_2.pdf}

    \includegraphics[width=\linewidth]
    {figures/arc1d/tasks/arc_1d_task_1d_pcopy_1c_46.pdf}

    \includegraphics[width=\linewidth]
    {figures/arc1d/tasks/arc_1d_task_1d_pcopy_mc_6.pdf}

    \caption{\textbf{ARC-1D denoising and pattern-copy tasks.}
    Examples of the denoising and pattern-copy task families, shown for both
    the single-color (1C) and multicolor (MC) variants.}
    \label{fig:arc1d-denoising-pcopy}
\end{figure}

\newpage
\section{Compositional Generalisation Examples}
\label{sec:compositional_examples}
\begin{figure}[H]
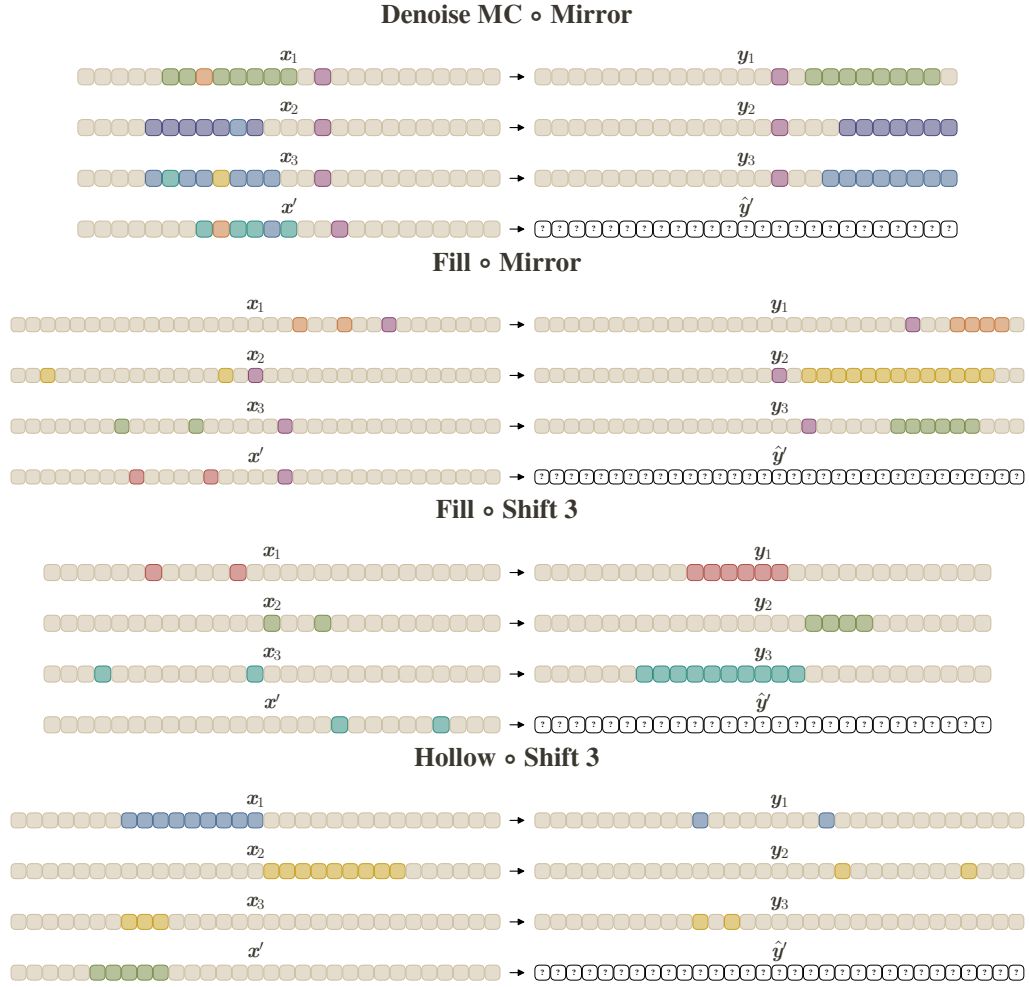

    \centering

    \includegraphics[width=\linewidth]{figures/arc1d/generalisation/compositional/arc_1d_task_1d_comp_denoisemc_mirror_0.pdf}

    \includegraphics[width=\linewidth]{figures/arc1d/generalisation/compositional/arc_1d_task_1d_comp_fill_mirror_0.pdf}


    \includegraphics[width=\linewidth]{figures/arc1d/generalisation/compositional/arc_1d_task_1d_comp_fill_shift3_0.pdf}

    \includegraphics[width=\linewidth]{figures/arc1d/generalisation/compositional/arc_1d_task_1d_comp_hollow_shift3_0.pdf}

    \caption{\textbf{ARC-1D compositional task examples.}
    Examples of held-out compositions formed by combining two
    transformations observed independently during training.}
    \label{fig:arc1d-compositional-examples-1}
\end{figure}

\newpage

\begin{figure}[H]
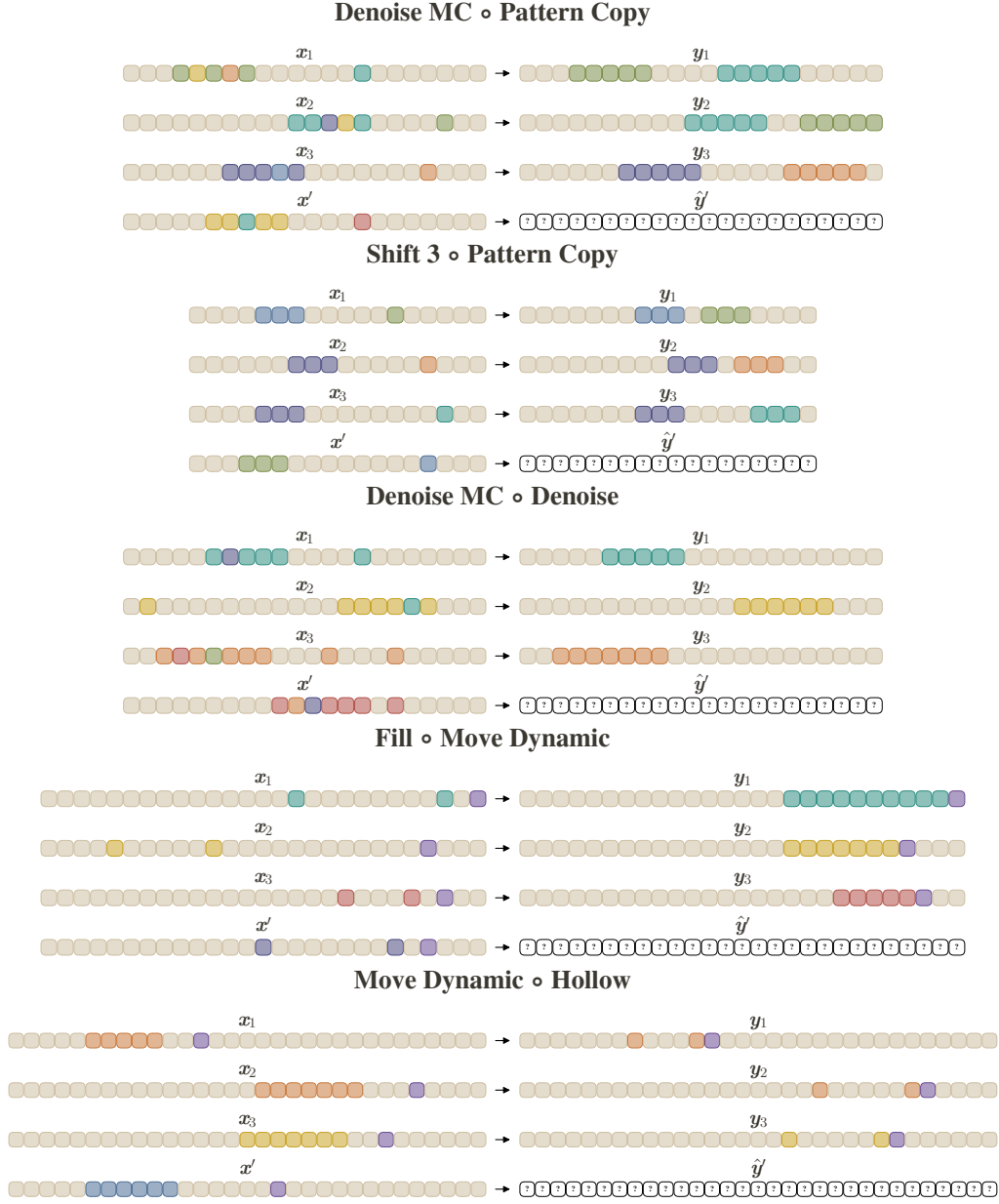

    \centering

    \includegraphics[width=\linewidth]{figures/arc1d/generalisation/compositional/arc_1d_task_1d_comp_denoisemc_copy_0.pdf}

    \includegraphics[width=\linewidth]{figures/arc1d/generalisation/compositional/arc_1d_task_1d_comp_shift3_copy_0.pdf}

    \includegraphics[width=\linewidth]{figures/arc1d/generalisation/compositional/arc_1d_task_1d_comp_denoisemc_denoise1c_0.pdf}

    \includegraphics[width=\linewidth]{figures/arc1d/generalisation/compositional/arc_1d_task_1d_comp_fill_movedynamic_0.pdf}

    \includegraphics[width=\linewidth]{figures/arc1d/generalisation/compositional/arc_1d_task_1d_comp_movedynamic_hollow_0.pdf}

    \caption{\textbf{ARC-1D compositional task examples.}
    Further examples of held-out compositions formed from transformations
    observed independently during training.}
    \label{fig:arc1d-compositional-examples-2}
\end{figure}


\end{document}